\documentclass[10pt,twocolumn,letterpaper]{article}

\usepackage[pagenumbers]{cvpr} 

\usepackage{booktabs}   
\usepackage{multirow}   
\usepackage{makecell}   
\usepackage{siunitx}    
\usepackage{caption}
\usepackage[table]{xcolor} 
\usepackage[pagebackref,breaklinks=true,colorlinks,citecolor=blue,urlcolor=blue,linkcolor=blue,bookmarks=false]{hyperref}

\usepackage{graphicx}
\usepackage{subcaption}

\newcommand{\blfootnote}[1]{\begingroup\renewcommand\thefootnote{}\footnote{#1}\addtocounter{footnote}{-1}\endgroup}

\def\paperID{4734} 
\def\confName{CVPR}
\def\confYear{2026}

\title{RecTok: Reconstruction Distillation along Rectified Flow}

\author{
Qingyu Shi$^{1,3*}$, Size Wu$^{2\dagger}$, Jinbin Bai$^{1}$, Kaidong Yu$^3$,
Yujing Wang$^1$, \\
Yunhai Tong$^{1\ddagger}$, Xiangtai Li$^{2}$, Xuelong Li$^{3}$ \vspace{5pt} \\
$^1$Peking University \quad $^2$Nanyang Technological University \quad $^3$TeleAI  \\
}

\begin{document}

\twocolumn[{
\maketitle
\begin{center}
    \captionsetup{hypcap=false}
    \includegraphics[width=\linewidth]{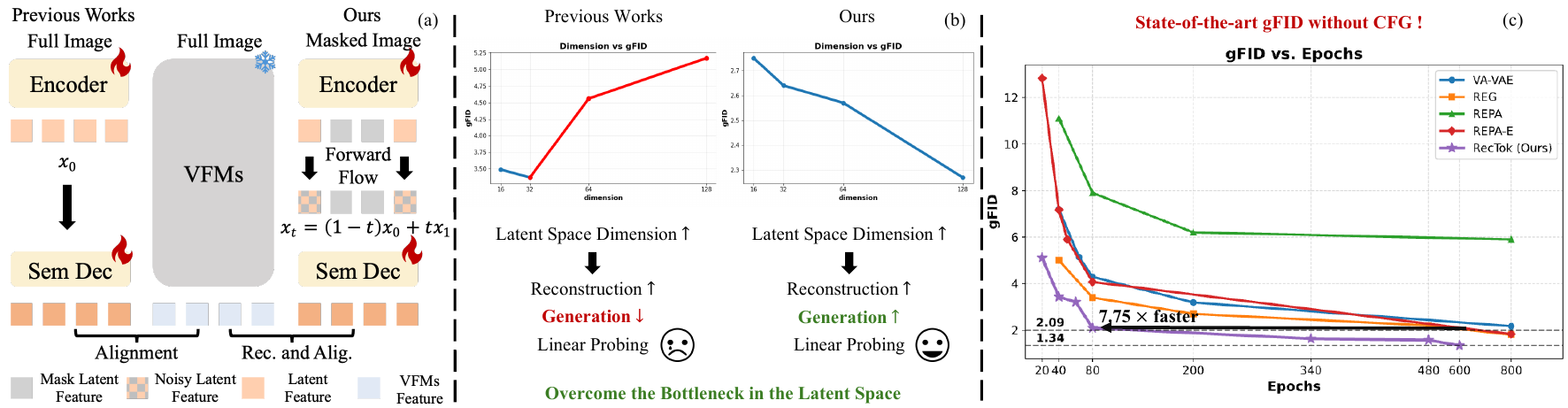}
    \captionof{figure}{(a) presents the core insights of our approach. Unlike previous works, we enhance semantic information along the forward pass of the rectified flow via reconstruction distillation. Fig. (b) shows that increasing the latent space dimension consistently improves the generation performance of \textbf{RecTok}, indicating that the dimensional bottleneck no longer constrains the semantic information encoded in the latent features. (c) compares the gFID convergence across training epochs, where our method converges \textbf{7.75$\times$ faster} than prior works and achieves a final gFID of \textbf{1.34 without classifier-free guidance}, the state-of-the-art gFID performance to date.}
    \label{fig:teaser}
\end{center}

\vspace{1em}
}]

\blfootnote{$^*$This work was completed at TeleAI. $^\dagger$Project lead. $^\ddagger$Corresponding author.}

\begin{abstract}
Visual tokenizers play a crucial role in diffusion models.
The dimensionality of latent space governs both reconstruction fidelity and the semantic expressiveness of the latent feature. 
However, a fundamental trade-off is inherent between dimensionality and generation quality, constraining existing methods to low-dimensional latent spaces. 
Although recent works have leveraged vision foundation models (VFMs) to enrich the semantics of visual tokenizers and accelerate convergence, high-dimensional tokenizers still underperform their low-dimensional counterparts.
In this work, we propose RecTok, which overcomes the limitations of high-dimensional visual tokenizers through two key innovations: flow semantic distillation and reconstruction–alignment distillation.
Our key insight is to make the forward flow in flow matching semantically rich, which serves as the training space of diffusion transformers, rather than focusing on the latent space as in previous works.
Specifically, our method distill the semantic information in VFMs into the forward flow trajectories in flow matching. And we further enhance the semantics by introducing a masked feature reconstruction loss.
Our RecTok achieves superior image reconstruction, generation quality, and discriminative performance. 
It achieves state-of-the-art results on the gFID-50K under both with and without classifier-free guidance settings, while maintaining a semantically rich latent space structure.
Furthermore, as the latent dimensionality increases, we observe consistent improvements.
Code and model are available at \url{https://shi-qingyu.github.io/rectok.github.io/}.
\end{abstract}    
\section{Introduction}
\label{sec:intro}

Diffusion modeling~\cite{ddpm, iddpm, vqdiffusion, LDM, SD3, flux} has become the dominant paradigm for image and video generation. 
As a crucial component, the visual tokenizer~\cite{vae, magvitv2} projects images from raw pixels to a compact latent space. 
Since the denoising network~\cite{dit, sit, uvit} is trained entirely in the latent space, computational cost is significantly reduced. 
However, the latent space is typically restricted to low feature dimensions to simplify diffusion training~\cite{LDM, SD3, flux}, which in turn limits both reconstruction fidelity and semantic expressiveness~\cite{ImprovDiffus, vavae}.
Therefore, expanding the latent space while maintaining generative stability becomes a fundamental challenge in training visual tokenizers.

To address this limitation, previous methods~\cite{vavae, maetok, gigatok} distill semantic information from vision foundation models (VFMs)~\cite{Dinov2, sam, MAE, siglip2} into the latent space, aiming to enrich representation capacity and accelerate generative convergence.
However, their generation quality in high dimensions still lags behind their low-dimensional counterparts.
Thus, these approaches remain constrained to low-dimensional latent spaces (\textit{e.g.}, dimension 32). 
Recently, RAE~\cite{RAE} increases DiT width to accommodate high-dimensional latents for diffusion training, achieving promising generative performance. 
However, its \textit{reconstruction performance} lags behind previous methods — a limitation that is detrimental to tasks such as editing~\cite{p2p,humanedit,feng2025item,viewcontrol,shi2025decouple} and personalized generation~\cite{dreambooth, DreamRelation}.
Furthermore, RAE does not systematically explore how dimensionality affects reconstruction, generation, and semantic representation. 
In this work, we revisit this question and present a principled framework for training high-dimensional visual tokenizers without compromising performance.

Unlike previous works~\cite{vavae, maetok, dinov2l} that directly inject semantics to the un-noised latent $x_0$, we take a more training-consistent perspective: Since DiT is trained on the forward flow $\{x_t \mid t \in [0,1]\}$ rather than on $x_0$, we enhance the semantics of all flow states $x_t$.
To understand the importance of semantic consistency along the flow, we first evaluate the discriminative capability of latent features across ${x_t}$. As shown in Fig.~\ref{fig:linearprobing}, the linear probing accuracy of several representative tokenizers~\cite{vavae, ldetok, vae} drops remarkably as the latent is propagated along the forward flow---the very representations that DiT receives during diffusion training.
This degradation highlights the need for semantic enhancement throughout the entire flow, not just at $x_0$.
In this work, we propose \emph{RecTok} with two key innovations to enhance semantic consistency along the forward flow, simultaneously improving dimensionality and generative quality.

The \textit{first} innovation is \textbf{Flow Semantic Distillation (FSD)}. Our key insight is to distill the semantics of VFMs into the forward flow trajectory \(\{x_t \mid t \in [0,1]\}\), which represents the interpolation of clean data \(x_0\) and noise \(x_1\). 
We utilize a lightweight semantic decoder to extract semantic features from points along the flow. These features are supervised by the corresponding representations from VFMs, as illustrated in Fig.~\ref{fig:teaser} (a). 
FSD explicitly encourages the forward flow path \(\{x_t \mid t \in [0,1]\}\) to remain semantically discriminative. Consequently, our RecTok exhibits even better accuracy on the flow than the latent features, as shown in Fig.~\ref{fig:linearprobing}.
The \textit{second} innovation is \textbf{Reconstruction and Alignment Distillation (RAD)}.
%
Inspired by masked image modeling methods~\cite{MAE,ibot, wei2022masked}, which obtain semantically rich features through pixel or feature reconstruction, we introduce a reconstructive target during FSD. 
Specifically, we apply random masks to the input image and reconstruct the missing regions based on the visible noisy latent features. 
We align the reconstructed latent features with full image features extracted from VFMs.

Following previous works~\cite{vavae, ldetok, RAE}, we train and evaluate our tokenizer and DiT~\cite{RAE} on the ImageNet-1K dataset~\cite{imagenet}. 
%
As the latent dimensionality increases, we observe consistent improvements across reconstruction, generation, and linear probing tasks. 
Compared to other distillation or VFM-based visual tokenizers, our approach exhibits a clear advantage in convergence speed and generation quality, especially under without classifier-free guidance~\cite{CFG} setting. 
%
%
To summarize, our key contributions include:
\begin{itemize}
    \item We identify the significance of enhancing semantics of forward flow trajectories, and introduce FSD and RAD that effectively expedite diffusion training.
    \item Our tokenizer achieves an effective balance among reconstruction, generation quality, and semantic representation.
    \item We demonstrate that all three aspects mentioned above can be consistently improved by increasing the dimensionality of the latent space.
\end{itemize}

\section{Related Work}
\label{sec:related_work}
\noindent
\textbf{Visual Tokenizers for Image Generation.}
%
%
Broadly, visual tokenizers fall into two categories: discrete and continuous. Discrete tokenizers quantize image features with a learnable codebook~\citep{vqvae, vqvae2, VQGAN, ViT-VQGAN}, and later works focus on enlarging the codebook and improving utilization~\citep{magvit, magvitv2}. Despite these advances, their reconstruction quality remains inferior to continuous tokenizers, limiting downstream generative performance~\citep{meissonic,muddit,lumina-dimoo,masks2worlds}.
Continuous tokenizers instead map images into a continuous latent space. Representative models such as VAE~\citep{vae} regularize this latent space using a KL loss, while subsequent works~\citep{VQGAN} introduce perceptual and adversarial losses to improve reconstruction quality. Recent studies~\citep{vavae, maetok} have also shown the advantage of aligning the latent space with the features of Vision Foundation Models (VFMs)~\citep{Dinov2, siglip2}, which accelerates convergence and improves downstream generation quality.
However, these approaches still restrict the latent representation to a low-dimensional space, constraining semantic expressiveness and reconstruction fidelity. 
In contrast, our work further expands the dimensionality of the latent space and observes continued improvements in generation quality.

\noindent
\textbf{High-dimensional Latent Space for Diffusion Models.}
%
A high-dimensional latent space is crucial for high-fidelity reconstruction and preserving rich semantic information. 
However, scaling latent dimensionality presents an inherent optimization challenge that often degrades generative performance.
Although VA-VAE~\citep{vavae} alleviates part of this difficulty via a VFM loss, its convergence in high dimensions remains noticeably slower than that of low-dimensional variants.
Another line of works~\cite{dinov2l, VFMTok} initialize the visual encoder with VFMs. Yet, these methods still project high-dimensional features into low-dimensional latents (e.g., dimension 32), inevitably discarding rich semantics in the VFMs. 
More recently, RAE~\cite{RAE} makes the first attempt to perform diffusion directly in the high-dimensional feature space of VFMs.
However, because the VFM is kept frozen, this approach inevitably loses fine-grained details, leading to reconstruction artifacts.
In concurrent work, SVG~\cite{svg, svgt2i} employs a residual encoder to enhance reconstruction fidelity while preserving semantics from VFMs. Nevertheless, a performance gap remains between SVG and state-of-the-art generation methods.
In this work, we develop a high-dimensional visual tokenizer that simultaneously excels in reconstruction fidelity, generative capability, and semantic representation.

\section{Method}
\label{sec:method}
%
%
%
\begin{figure}[t]
  \centering
  \begin{subfigure}{0.48\linewidth}
    \centering
    \includegraphics[width=\linewidth]{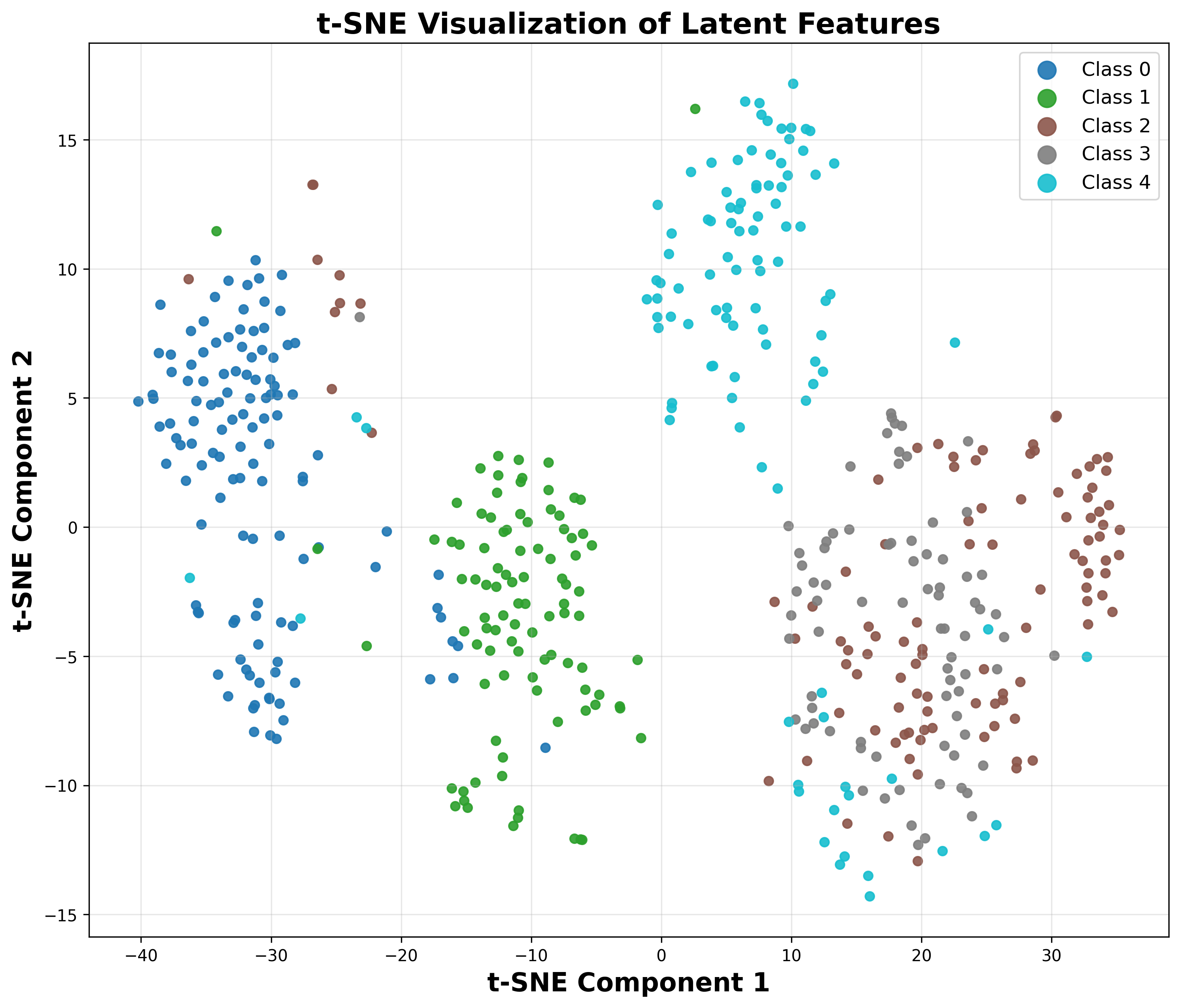}
    \caption{RecTok, L.P. Acc.=55.61\%}
    \label{fig:grid:a}
  \end{subfigure}\hfill
  \begin{subfigure}{0.48\linewidth}
    \centering
    \includegraphics[width=\linewidth]{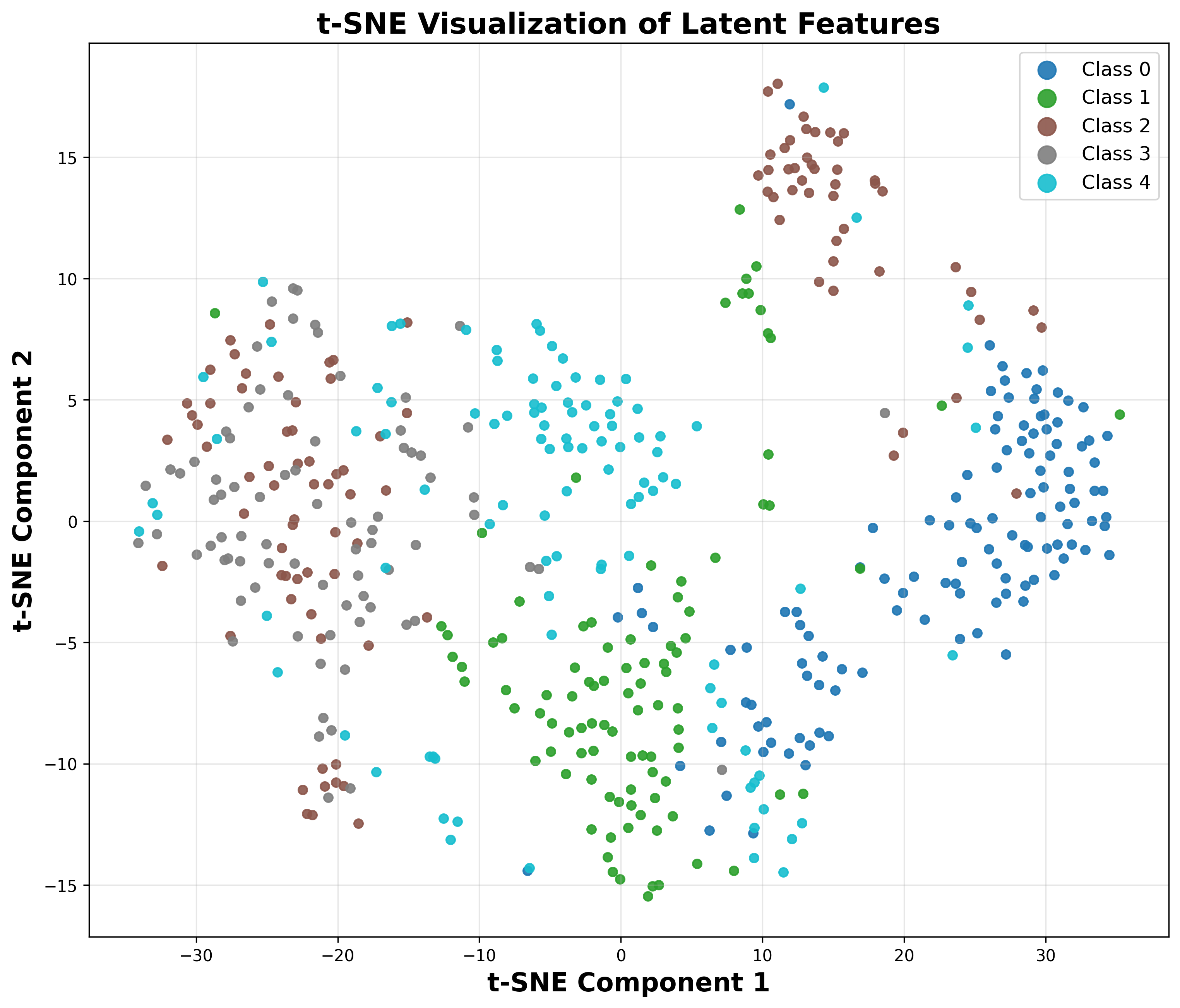}
    \caption{VA-VAE, L.P. Acc.=18.36\%}
    \label{fig:grid:b}
  \end{subfigure}

  \vspace{0.5em} 

  \begin{subfigure}{0.48\linewidth}
    \centering
    \includegraphics[width=\linewidth]{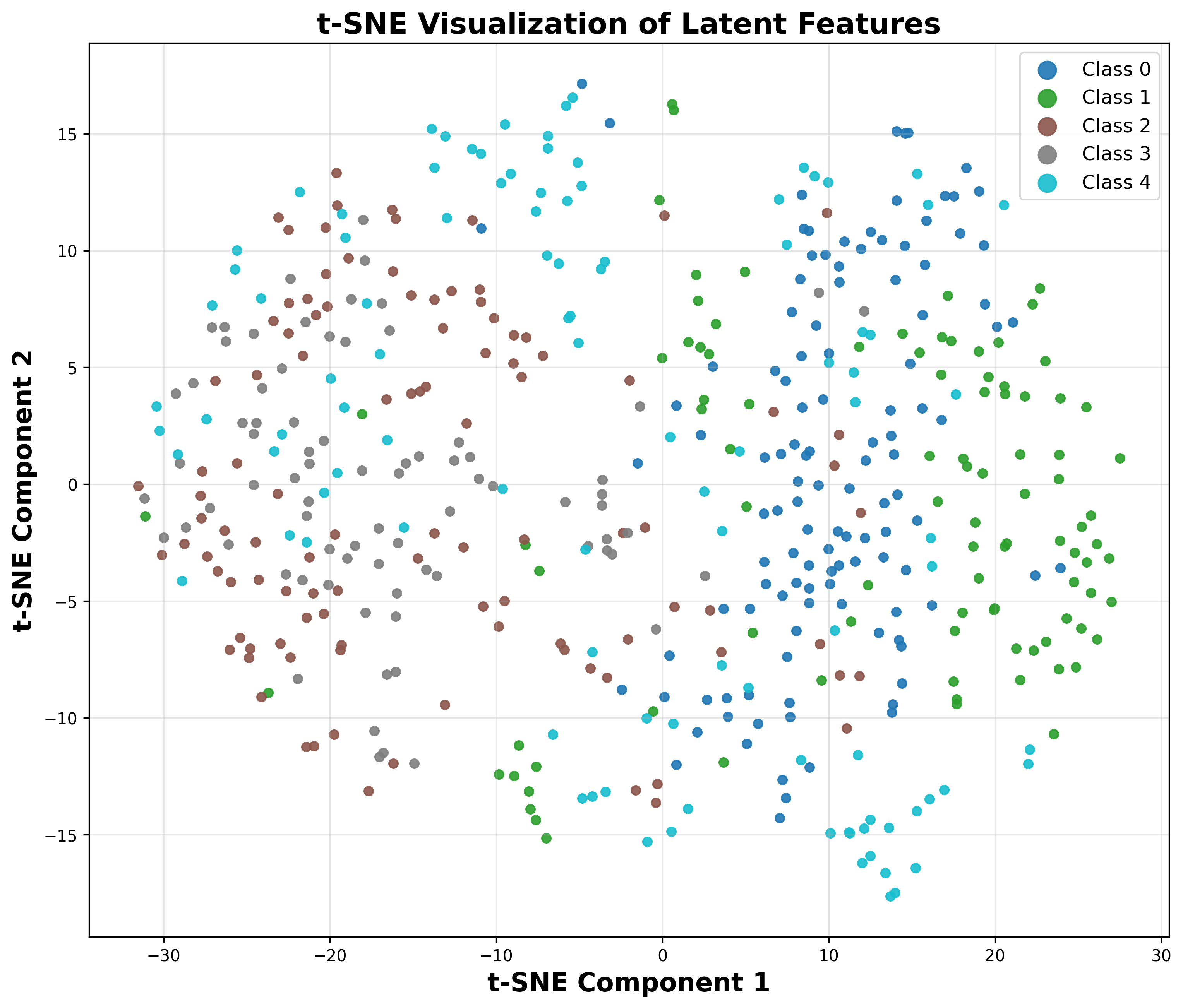}
    \caption{DeTok, L.P. Acc.=10.2\%}
    \label{fig:grid:c}
  \end{subfigure}\hfill
  \begin{subfigure}{0.48\linewidth}
    \centering
    \includegraphics[width=\linewidth]{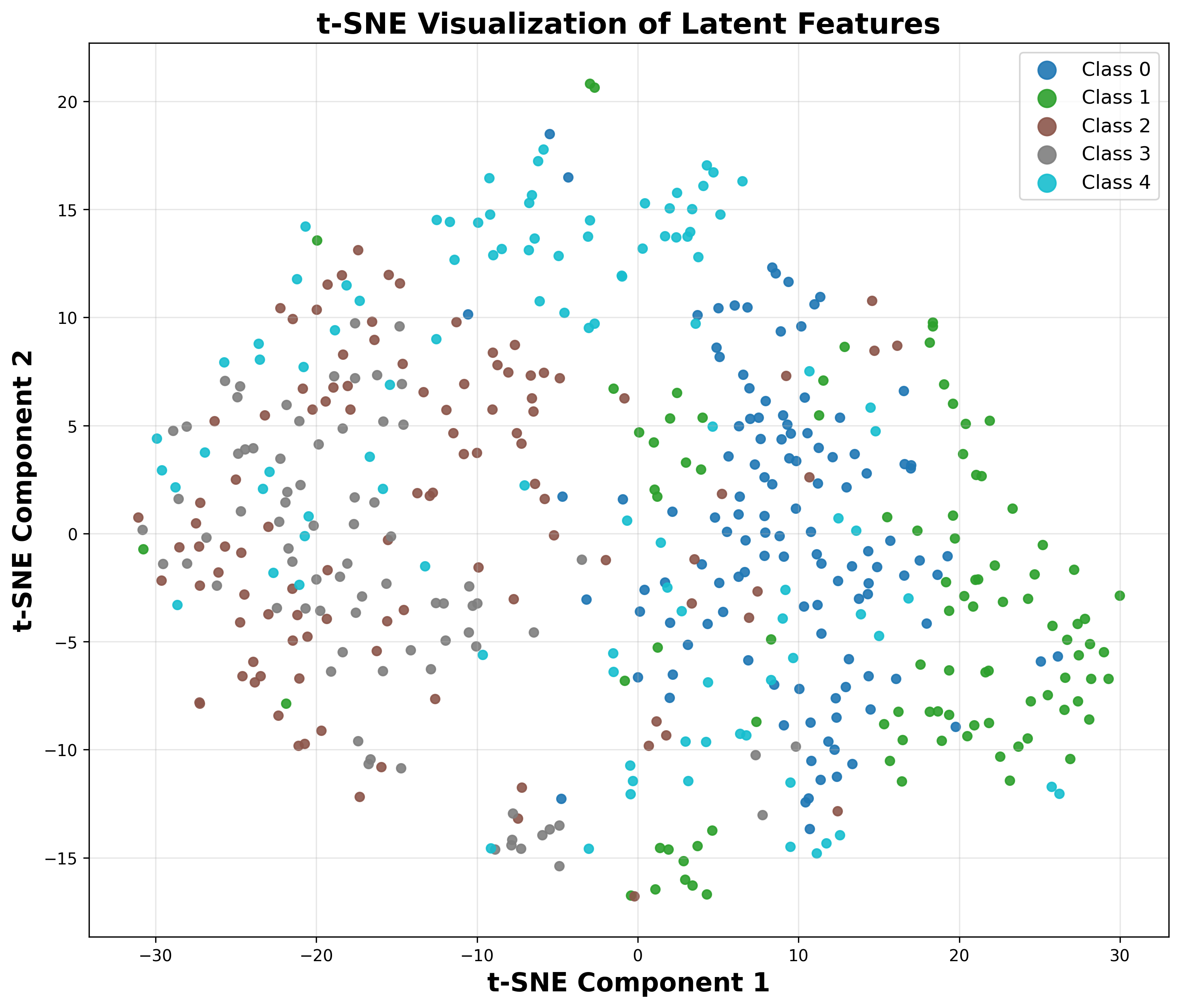}
    \caption{VAE, L.P. Acc.=5.4\%}
    \label{fig:grid:d}
  \end{subfigure}

  \caption{\textbf{Linear probing results on $x_t$.} We evaluate the discriminative ability of representative tokenizers on the forward flow through linear probing on $x_t$. Specifically, we fix $t = 0.5$. As shown in Fig.~\ref{fig:grid:b}--\ref{fig:grid:d}, both the t-SNE visualization and the linear probing accuracy demonstrate that their latent features perform poorly during the training of DiT. In contrast, our RecTok exhibits a clear advantage even under noise interpolation.}
  \label{fig:linearprobing}
\end{figure}

\subsection{Rectified Flow in Image Generation}

\noindent
\textbf{Flow Matching.}
Flow matching methods~\cite{flowmatching} construct a distribution transformation between data $x_0$ and noise $x_1$ through forward and reverse flows. 
As a representative approach, Rectified Flow~\cite{rf} adopts a forward flow defined by linear interpolation, which simplifies the formulation of the velocity field:
\begin{equation}
    x_t = (1 - t)\, x_0 + t\, x_1, \quad v_t = \frac{d x_t}{d t} = x_1 - x_0.
    \label{eq:rf_interp}
\end{equation}

During training, a neural network $v_\theta(x, t)$ is optimized on the forward flow $\{x_t\mid t\in(0, 1]\}$ to approximate the velocity field as:
\begin{equation}
    \mathcal{L}_{\text{RF}}
    \;=\;
    \mathbb{E}_{\,t,\, x_0,\, x_1}
    \left[
        \big\|\, v_\theta(x_t, t) - (x_1 - x_0) \,\big\|_2^2
    \right],
    \label{eq:rf_loss}
\end{equation}

During generation, $v_\theta(x,t)$ predicts the velocity that gradually transforms noisy data into clean data $x_0$ through the ODE solver~\cite{rf, dpmsolver}.

\noindent
\textbf{Visual Tokenizers.}
To reduce the training cost of generation models, prior works~\cite{LDM, SD3, flux} project images into a compact latent space via an encoder-decoder image tokenizer~\cite{vae}. Given an input image \(I \in \mathbb{R}^{H \times W \times 3}\), the encoder \(E_\theta\) produces a latent \(x_0 \in \mathbb{R}^{h \times w \times c}\), and the decoder \(D_\phi\) reconstructs \(\hat{I}\):
\begin{equation}
x_0 = E_\theta(I), 
\qquad 
\hat{I} = D_\phi(x_0).
\label{eq:encdec_det}
\end{equation}

Both the encoder and decoder are typically based on CNN~\cite{cnn} or ViT~\cite{vit} architectures. Considering computational efficiency and scalability~\cite{RAE}, we adopt a ViT-based encoder–decoder design in this work.

The training of visual tokenizers typically involves multiple objectives, including reconstruction loss, perceptual loss, GAN loss~\cite{vqvae}, and KL loss~\cite{vae}. 
Moreover, recent studies~\cite{vavae} have shown that distilling semantic information from Vision Foundation Models (VFMs)~\cite{Dinov2, MAE} into the latent space can accelerate the convergence of downstream generative models and further improve image quality. 
Overall, the general loss function of the visual tokenizer can be formulated as follows:
\begin{equation}
\mathcal{L}
=
\lambda_{\text{rec}}\,\mathcal{L}_{\text{rec}}
+
\lambda_{\text{per}}\,\mathcal{L}_{\text{per}}
+
\lambda_{\text{GAN}}\,\mathcal{L}_{\text{GAN}}
+
\lambda_{\text{KL}}\,\mathcal{L}_{\text{KL}}
+
\lambda_{\text{sem}}\,\mathcal{L}_{\text{sem}}.
\label{eq:total}
\end{equation}

\begin{figure*}[htbp]
    \centering
    \includegraphics[width=1.0\linewidth]{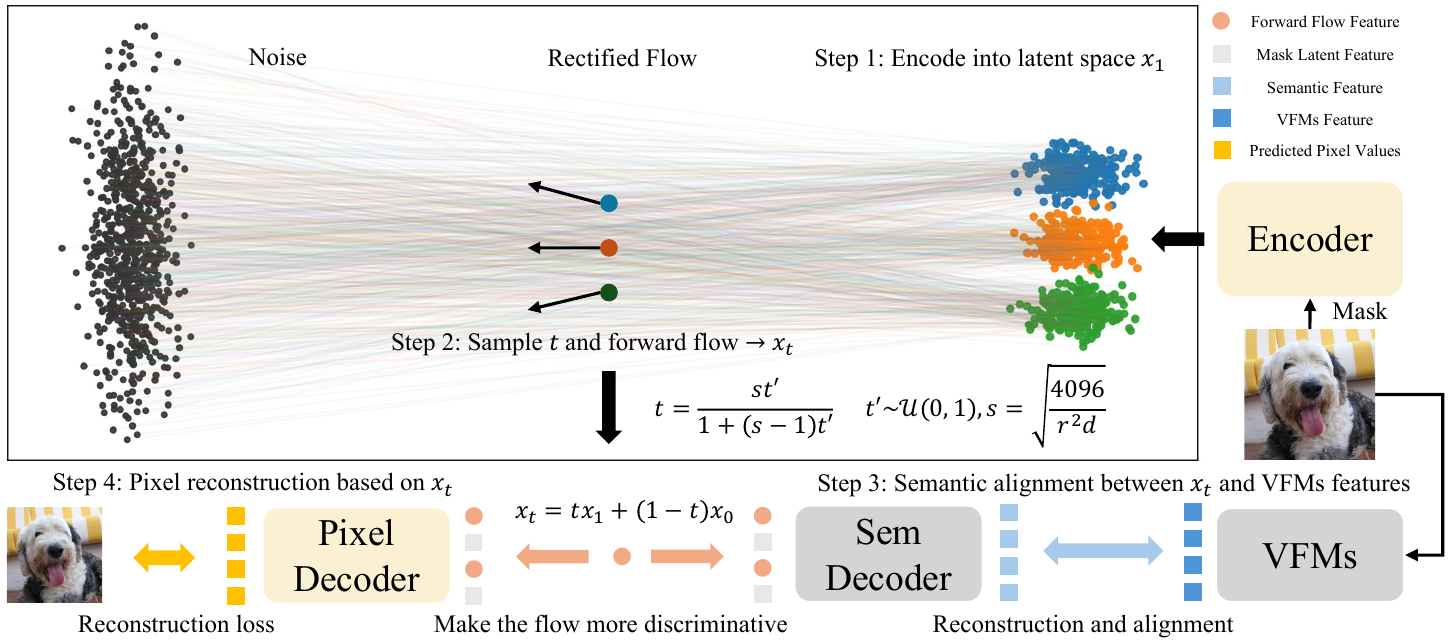}
    \caption{\textbf{Pipeline of RecTok.} During the training of RecTok, we apply a random mask to the input image and encode the visible regions using the encoder to obtain $x_1$. We then sample a time step $t$ and use the forward flow to generate the corresponding $x_t$. Subsequently, $x_t$ is fed into two decoders: the Semantic Decoder reconstructs the features of VFMs, while the Pixel Decoder reconstructs the pixel space. After training, both the Semantic Decoder and VFMs are discarded, ensuring the efficiency of RecTok during inference.}
    \label{fig:pipeline}
\end{figure*}

\begin{table*}[t]
\centering
\small
\renewcommand{\arraystretch}{1.1}
\setlength{\tabcolsep}{5pt} 
\makebox[\textwidth][c]{
\resizebox{0.9\textwidth}{!}{
\begin{tabular}{l c c cccc cccc}
\toprule
\multirow{2}{*}{\textbf{Method}} & \multirow{2}{*}{\makecell{\textbf{Epochs}}} & \multirow{2}{*}{\textbf{Params}} & \multicolumn{4}{c}{\textbf{Generation@256 w/o guidance}} & \multicolumn{4}{c}{\textbf{Generation@256 w/ guidance}} \\
\cmidrule(lr){4-7} \cmidrule(lr){8-11}
 & & & \textbf{gFID}$\downarrow$ & \textbf{IS}$\uparrow$ & \textbf{Prec.}$\uparrow$ & \textbf{Rec.}$\uparrow$ & \textbf{gFID}$\downarrow$ & \textbf{IS}$\uparrow$ & \textbf{Prec.}$\uparrow$ & \textbf{Rec.}$\uparrow$ \\
\midrule
\multicolumn{11}{l}{\textit{\textbf{Autoregressive}}} \\
\arrayrulecolor{black!30}\midrule
\textcolor{black!40}{VAR~\citep{VAR}} & 
\textcolor{black!40}{350} & 
\textcolor{black!40}{2.0B} & 
\textcolor{black!40}{1.92} & 
\textcolor{black!40}{\textbf{323.1}} &
\textcolor{black!40}{\textbf{0.82}} & 
\textcolor{black!40}{0.59} & 
\textcolor{black!40}{1.73} & 
\textcolor{black!40}{\textbf{350.2}} & 
\textcolor{black!40}{0.82} & 
\textcolor{black!40}{0.60} \\
\textcolor{black!40}{MAR~\citep{mar}} & 
\textcolor{black!40}{800} & 
\textcolor{black!40}{943M} & 
\textcolor{black!40}{2.35} & 
\textcolor{black!40}{227.8} & 
\textcolor{black!40}{0.79} & 
\textcolor{black!40}{0.62} & 
\textcolor{black!40}{1.55} & 
\textcolor{black!40}{303.7} & 
\textcolor{black!40}{0.81} & 
\textcolor{black!40}{0.62} \\
\textcolor{black!40}{$l$-DeTok~\citep{ldetok}} & 
\textcolor{black!40}{800} & 
\textcolor{black!40}{479M} & 
\textcolor{black!40}{1.86} & 
\textcolor{black!40}{238.6} & 
\textcolor{black!40}{\textbf{0.82}} & 
\textcolor{black!40}{0.61} & 
\textcolor{black!40}{1.35} & 
\textcolor{black!40}{304.1} & 
\textcolor{black!40}{0.81} & 
\textcolor{black!40}{0.62} \\
\arrayrulecolor{black}\midrule
\multicolumn{11}{l}{\textit{\textbf{Pixel Diffusion}}} \\
\arrayrulecolor{black!30}\midrule
ADM~\citep{adm} &  400  &  554M  & 10.94 &  101.0 & 0.69 & 0.63 & 3.94 & 215.8 & \textbf{0.83} & 0.53\\
RIN~\citep{rin} &  480  &  410M  & 3.42  & 182.0  &  -   &  -   &  -   &   -    &  -   &  -  \\
PixelFlow~\citep{pixelflow} & 320 & 677M & - & -   &   -  &  -   & 1.98 & 282.1 & 0.81 & 0.60 \\
PixNerd~\citep{pixnerd} & 160 & 700M & -   &  -       &  - &  -  &  2.15 & 297.0 & 0.79 & 0.59 \\
JiT~\citep{jit} & 600 & 2.0B & - &  -  &  - &  -  &  1.82 & 292.6 & - & - \\
\arrayrulecolor{black}\midrule
\multicolumn{11}{l}{\textit{\textbf{Latent Diffusion}}} \\
\arrayrulecolor{black!30}\midrule
DiT~\citep{dit} & 1400 & 675M         & 9.62 & 121.5 & 0.67 & 0.67 & 2.27 & 278.2 & \textbf{0.83} & 0.57 \\
MaskDiT~\citep{maskdit} & 1600 & 675M & 5.69 & 177.9 & 0.74 & 0.60 & 2.28 & 276.6 & 0.80 & 0.61 \\
SiT~\citep{sit} & 1400 & 675M         & 8.61 & 131.7 & 0.68 & 0.67 & 2.06 & 270.3 & 0.82 & 0.59 \\
MDTv2~\citep{MDTv2} & 1080 & 675M & - & - & - & - & 1.58 & 314.7 & 0.79 & 0.65 \\
\midrule
\multirow{2}{*}{VA-VAE~\citep{vavae}} & 80 & \multirow{2}{*}{675M} & 4.29 & - & - & - & - & - & - & -  \\
 & 800 &  & 2.17 & 205.6 & 0.77 & 0.65 & 1.35 & 295.3 & 0.79 & 0.65 \\
\midrule
AFM~\citep{dinov2l} & 800 & 675M & 2.04 & 206.2 & 0.76 & 0.67 & 1.37 & 293.6 & 0.79 & 0.65  \\
\midrule
\multirow{2}{*}{REPA~\citep{repa}} & 80 & \multirow{2}{*}{675M} & 7.94 & 121.3 & 0.69 & 0.64 & - & - & - & - \\
                                   & 800 &                      & 5.90 & 157.8 & 0.70 & \textbf{0.69} & 1.42 & 305.7 & 0.80 & 0.64 \\
\midrule
\multirow{2}{*}{DDT~\citep{ddt}} & 80 & \multirow{2}{*}{675M} & 6.62 & 135.2 & 0.69 & 0.67 & 1.52 & 263.7 & 0.78 & 0.63 \\
& 400 &  & 6.27 & 154.7 & 0.68 & \textbf{0.69} & 1.26 & 310.6 & 0.79 & 0.65\\
\midrule
\multirow{2}{*}{REPA-E~\citep{repa-e}} & 80 & \multirow{2}{*}{675M} & 3.46 & 159.8 & 0.77 & 0.63 & 1.67 & 266.3 & 0.80 & 0.63 \\
 & 800 &  & 1.83 & 217.3 & 0.77 & 0.66 & 1.26 & \textbf{314.9} & 0.79 & 0.66 \\
\midrule
SVGTok~\citep{svg} & 1400 & 675M & 3.36 & 181.2 & - & - & 1.92 & 264.9 & - & - \\
\midrule
\multirow{2}{*}{RAE~\citep{RAE}} & 80 & \multirow{2}{*}{839M} & 2.16 & 214.8 & \textbf{0.82} & 0.59 & -- & -- & -- & -- \\
& 800 & & 1.51 & 242.9 & 0.79 & 0.63 & \textbf{1.13} & 262.6 & 0.78 & \textbf{0.67} \\
\arrayrulecolor{black!30}\midrule
\multirow{2}{*}{RecTok (Ours)} & 80 & \multirow{2}{*}{839M} & 2.09 & 198.6 & 0.79 & 0.62 & 1.48 & 223.8 & 0.79 & 0.65 \\
& 600 & & \textbf{1.34} & \textbf{254.6} & 0.78 & 0.65 & \textbf{1.13} & 289.2 & 0.79 & \textbf{0.67} \\
\arrayrulecolor{black}\bottomrule
\end{tabular}
}
}
\caption{\textbf{Class-conditional performance on ImageNet 256$\times$256.} 
RecTok reaches an FID of 1.34 and an IS of 254.6 without guidance, outperforming previous methods by a large margin. 
With AutoGuidance~\cite{AG}, it achieves an FID of 1.13 and an IS of 289.2 using only 600 epochs, representing the best overall performance.
}
\label{tab:main_res}
\end{table*}

\subsection{Our Method: RecTok}

Our motivation is illustrated in Fig.~\ref{fig:linearprobing}. 
Although previous methods improve the semantic information in $x_0$, the discriminative ability of $x_t$ deteriorates significantly when training the diffusion transformers (DiTs). 
Our key insight is to enhance the semantic information not only in $x_0$, but also the forward flow $\{x_t\mid t\in[0, 1]\}$, where the DiTs are trained.
In the following section, we present two key innovations that enhance semantic representation throughout the forward flow.

\noindent
\textbf{Flow Semantic Distillation (FSD).}
Our goal is to make every point $x_t$ along the forward flow discriminative and semantically rich. 
%
Fortunately, the forward flow from data $x_0$ to noise $\epsilon$ is independent of the velocity network $v_\theta(x,t)$, allowing us to obtain $x_t=(1-t)x_0+t\epsilon, \ t\in [0, 1]$ easily through interpolation between the encoded $x_0 = E_{\theta}(I)$ and Gaussian noise $\epsilon$. 
Each $x_t$ is then decoded by a lightweight semantic decoder $D_{\text{sem}}$ to obtain semantic features, which are supervised by Vision Foundation Models (VFMs) $E_{\text{VFM}}$:
\begin{equation}
    \mathcal{L}_{\text{sem}} = 1 - \text{cos}(D_{\text{sem}}(x_t), E_{\text{VFM}}(I))
\end{equation}

Specifically, the lightweight semantic decoder $D_{\text{sem}}$ adopts a transformer architecture with only 1.5M parameters. A lightweight design enforces the encoder to capture richer semantic representations, as an overly powerful semantic decoder would otherwise draw away the semantic information from the encoder.
%
We remove the normalization on $D_{\text{sem}}(x_t)$ and $E_{\text{VFM}}(I)$ for simplicity.

During FSD, we need to sample the timestep t. 
Considering the redundancy in high-dimensional latent spaces, we apply a dimension-dependent shift to the distribution of t, following RAE~\cite{RAE}, and sample it as follows:
\begin{equation}
t = \frac{s t'}{1 + (s - 1)t'}, \quad t' \sim \mathcal{U}(0,1), \quad s = \sqrt{\frac{4096}{r^2 d}}
\end{equation}
where $r, d$ is the resolution and dimension of the latent feature, respectively.

\noindent
\textbf{Reconstruction and Alignment Distillation (RAD).}
Inspired by masked image modeling methods~\cite{MAE,ldetok,maetok}, which enforce the model to learn robust representations by predicting unseen image patches. 
To further enhance the semantics along the flow. We introduce a reconstruction target in the FSD. 
Specifically, we apply random masks to the input image. We use a random mask ratio between -0.1 and 0.4. A negative ratio means that no mask is applied.
After encoding the visible image into latent feature $x_0^{\text{vis}}$, we utilize a semantic decoder to reconstruct VFM features based on the $x_t^{\text{vis}}=(1-t)x_0^{\text{vis}}+t\epsilon$.
To ensure compatibility with the reconstruction task, we utilize a transformer-based semantic decoder $D_{sem}$.
The semantic loss \(\mathcal{L}_{\text{sem}}\) is applied to both masked and unmasked regions. 
Our ablation study demonstrates that jointly performing semantic alignment and reconstruction yields the best overall performance.

\noindent
\textbf{Dimension of Latent Space.}
A fundamental limitation of previous tokenizers in generative models is their confinement to low-dimensional latent spaces. 
Although semantic distillation~\cite{vavae} and channel regularization~\cite{dcae15} partially alleviate this issue, the best practice remains restricted to 32 dimensions. 
We progressively increase the dimensionality of the latent space. 
As shown in Tab.~\ref{tab:dim-results}. Interestingly, this leads to consistent improvements in reconstruction (rFID, PSNR), generation (gFID, IS), and semantics (linear probing). 
This finding suggests the emergence of a shared latent space in higher dimensions that effectively supports low-level and high-level tasks.

\noindent
\textbf{Decoder Finetuning.}
After joint pixel and VFM-feature training, we freeze the encoder to preserve the learned latent semantics and finetune only the pixel decoder for image reconstruction.
We disable the FSD and RAD and remove the losses $\mathcal{L}_{\text{KL}}$ and $\mathcal{L}_{\text{sem}}$.
While we do not claim this as our primary contribution, it is a crucial step to improve reliability and quality of the reconstruction. We show the performance improvement in Tab.~\ref{tab:overall_ablation}.
\section{Experiment}
\label{sec:exp}

\subsection{Implementation Details}

\noindent
\textbf{Visual Tokenizer.}
We adopt an architecture and training strategy largely following $l$-DeTok~\citep{ldetok}. Specifically, we employ a ViT-B~\citep{vit} backbone equipped with ROPE~\citep{rope}, SwiGLU~\cite{swiglu}, and RMSNorm~\cite{rmsnorm} for both encoder and decoder.
To investigate the impact of dimension on semantics, generation, and reconstruction, we train models with latent dimensions of 16, 32, 64, and 128. Note that this only affects the dimensionality of the ViT’s linear head, so the resulting changes in parameter count and computational cost are negligible. Reparameterization and KL divergence are used to regularize the latent space.
We train our tokenizer on the ImageNet-1K training set for 200 epochs. We set $\lambda_{rec}=1.0$, $\lambda_{per}=1.0$, $\lambda_{adv}=0.5$, $\lambda_{kl}=1\times 10^{-6}$, and $\lambda_{sem}=1$. The learning rate is set to $4\times 10^{-4}$, with a linear warmup during the first 50 epochs followed by a cosine decay schedule for the remaining 150 epochs. We use a global batch size of 1024 and an EMA rate of 0.999. We evaluate our tokenizer through rFID~\cite{fid} and PSNR on the ImageNet-1K validation set.

\noindent
\textbf{Diffusion Model.}
For the diffusion model, inspired by the advanced architecture of $\text{DiT}^{\text{DH}}$~\cite{RAE}, we utilize $\text{DiT}^{\text{DH}}\text{-XL}$ as our diffusion transformers. We train $\text{DiT}^{\text{DH}}\text{-XL}$ on ImageNet-1K using rectified flow with a timestep shift strategy. The model is trained for 800 epochs with an initial learning rate of $2\times 10^{-4}$ and global batch size 1024, followed by a linear decay to $2\times 10^{-5}$ after 40 epochs. During training, we apply gradient clipping with a value of 1.0 and no weight decay. We use an EMA rate of 0.995, and all evaluations are conducted using the EMA-weighted model. For the ablation studies, the model trains for 80 epochs using the same training strategy. We evaluate the diffusion models on the ImageNet-1K validation set, measuring gFID, Inception Score (IS)~\cite{is}, Precision, and Recall. We utilize AutoGuidance~\cite{AG} as the classifier-free guidance method. The bad version in the AutoGuidance is a $\text{DiT}^{\text{DH}}\text{-S}$ trained on ImageNet-1K for 30 epochs. During inference, we sample 150 steps using the Euler solver with a timestep shift; In the ablation studies, we sample only 50 steps and skip decoder finetuning to reduce computational cost.

All experiments are conducted on 32 H100 GPUs. Training the RecTok requires roughly 19 hours, while $\text{DiT}^{\text{DH}}$ requires 10 hours for 80 epochs and 3 days for 600 epochs.

\begin{table}[t]
\centering
\caption{
\textbf{Results across different feature dimensions.}
L.P. Acc. (L) denotes linear probing accuracy on latent features, 
while L.P. Acc. (SL) refers to linear probing accuracy on second-last layer features. 
As the feature dimension increases, discriminative ability, reconstruction, and generation show consistent gains.
}
\resizebox{1.0\linewidth}{!}{
\begin{tabular}{rrrrrr}
\toprule
\textbf{Dim} & \textbf{L.P. Acc. (L)} & \textbf{L.P. Acc. (SL)} & \textbf{rFID} & \textbf{PSNR} & \textbf{gFID} \\
\midrule
16  & 24.1 & 62.9 & 0.74 & 22.75 & 2.75 \\
32  & 38.8 & 63.7 & 0.71 & 24.08 & 2.64 \\
64  & 47.2 & 65.0 & 0.66 & 24.93 & 2.57 \\
\rowcolor{black!6}
128 & \textbf{55.4} & \textbf{68.1} & \textbf{0.65} & \textbf{25.28} & \textbf{2.27} \\
\bottomrule
\end{tabular}
}
\label{tab:dim-results}
\end{table}

\begin{table}[t]
\centering
\setlength{\tabcolsep}{8pt}
\caption{
\textbf{Tokenizer comparison on ImageNet-1K.}
We compare RecTok with representative tokenizers in terms of parameters, GFLOPs, reconstruction, and generation. 
RecTok achieves the best performance among ViT-based tokenizers.
}
\small 
\resizebox{1.0\linewidth}{!}{
\begin{tabular}{lccccc}
\toprule
\multirow{2}{*}{\textbf{Tokenizer}} & \multirow{2}{*}{\textbf{Params}} & \multirow{2}{*}{\textbf{GFlops}} & \multicolumn{3}{c}{\textbf{ImageNet}} \\
\cmidrule(lr){4-6}
& & & \textbf{rFID} & \textbf{PSNR} & \textbf{gFID} \\
\midrule
SD-VAE$^\dagger$  & 84M  & 445 & 0.62 & 26.04 & 8.30 \\
VA-VAE            & \textbf{70M}  & 310 & \textbf{0.28} & \textbf{26.30} & 2.17 \\
MAETok            & 176M & 54.2 & 0.48 & 23.61 & 2.21 \\
DeTok             & 176M & \textbf{44.4} & 0.52 & 23.53 & 1.86 \\  
RAE               & 395M & 128.9 & 0.57 & 18.98 & 1.51 \\
\rowcolor{black!6}
RecTok            & 176M & \textbf{44.4} & 0.48 & 26.16 & \textbf{1.34} \\
\bottomrule
\end{tabular}
}
\label{tab:tokenizer_comparison}
\vspace{2pt}
\footnotesize{$^\dagger$Numbers reported from the original papers.}
\end{table}

\begin{figure}[t]
    \centering
    \includegraphics[width=1.0\linewidth]{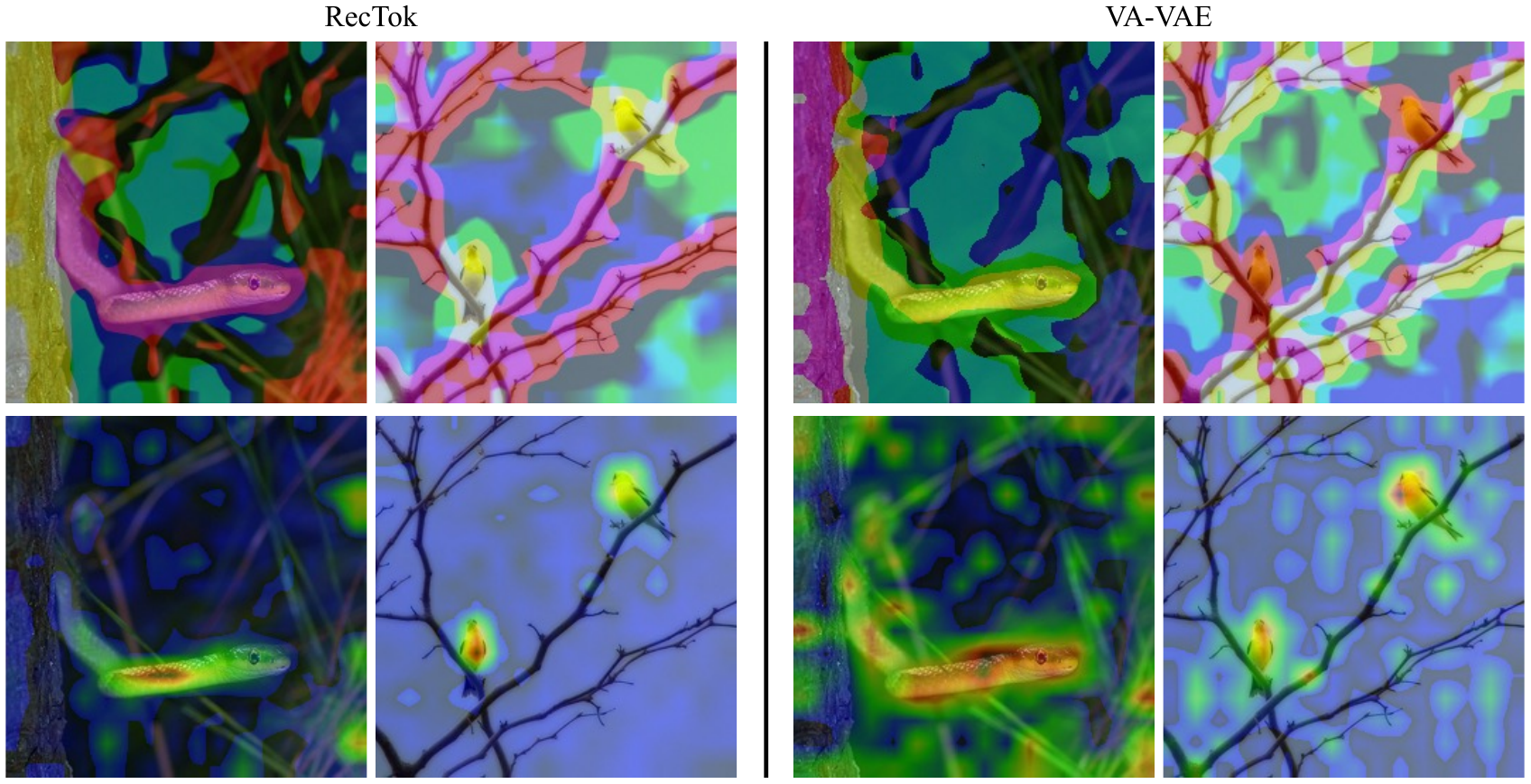}
    \caption{
    \textbf{Visualization of latent features.}
    We present the PCA projection and cosine similarity heatmap of the latent features from RecTok and VA-VAE. 
    RecTok exhibits a more semantically rich latent space.
    }
    \label{fig:feature_vis}
\end{figure}

\begin{figure*}[t]
    \centering
    \includegraphics[width=1.0\linewidth]{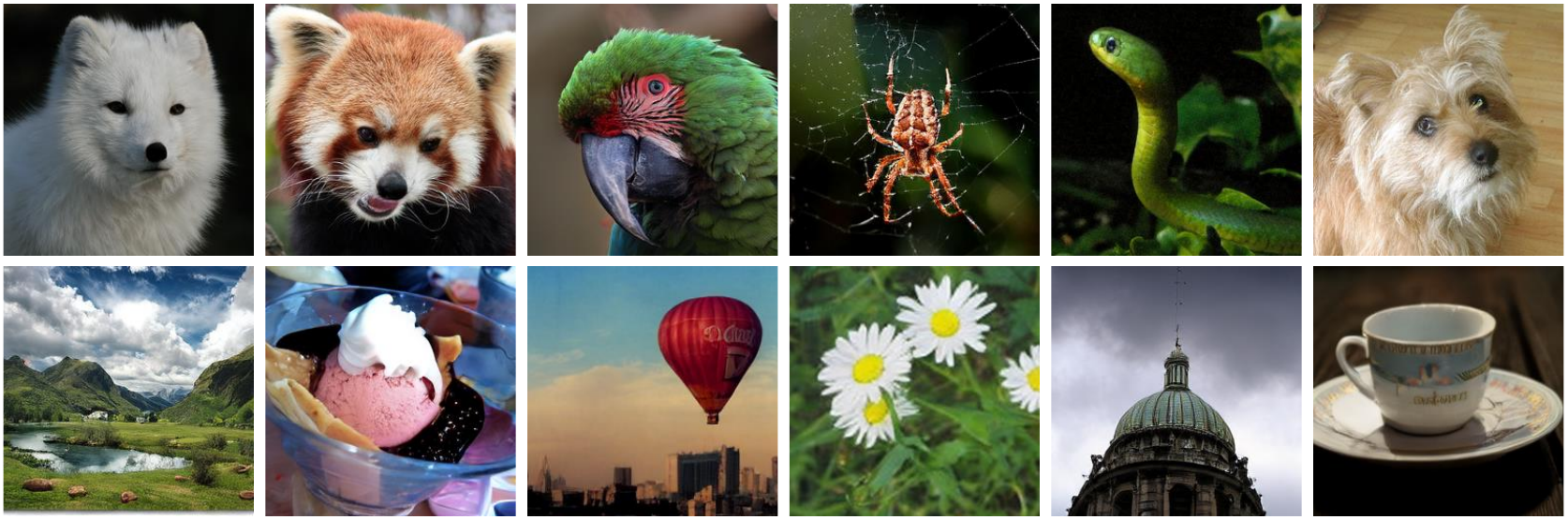}
    \caption{\textbf{Qualitative results on ImageNet-1K $\text{256}\times\text{256}$.}
    We show selected examples of class-conditional generation using $\text{DiT}^{\text{DH}}\text{-XL}$ with AutoGuidance.
    }
    \label{fig:qualitative_results}
\end{figure*}

\begin{figure}[t]
    \centering
    \begin{subfigure}[t]{0.48\linewidth}
        \centering
        \includegraphics[width=\linewidth]{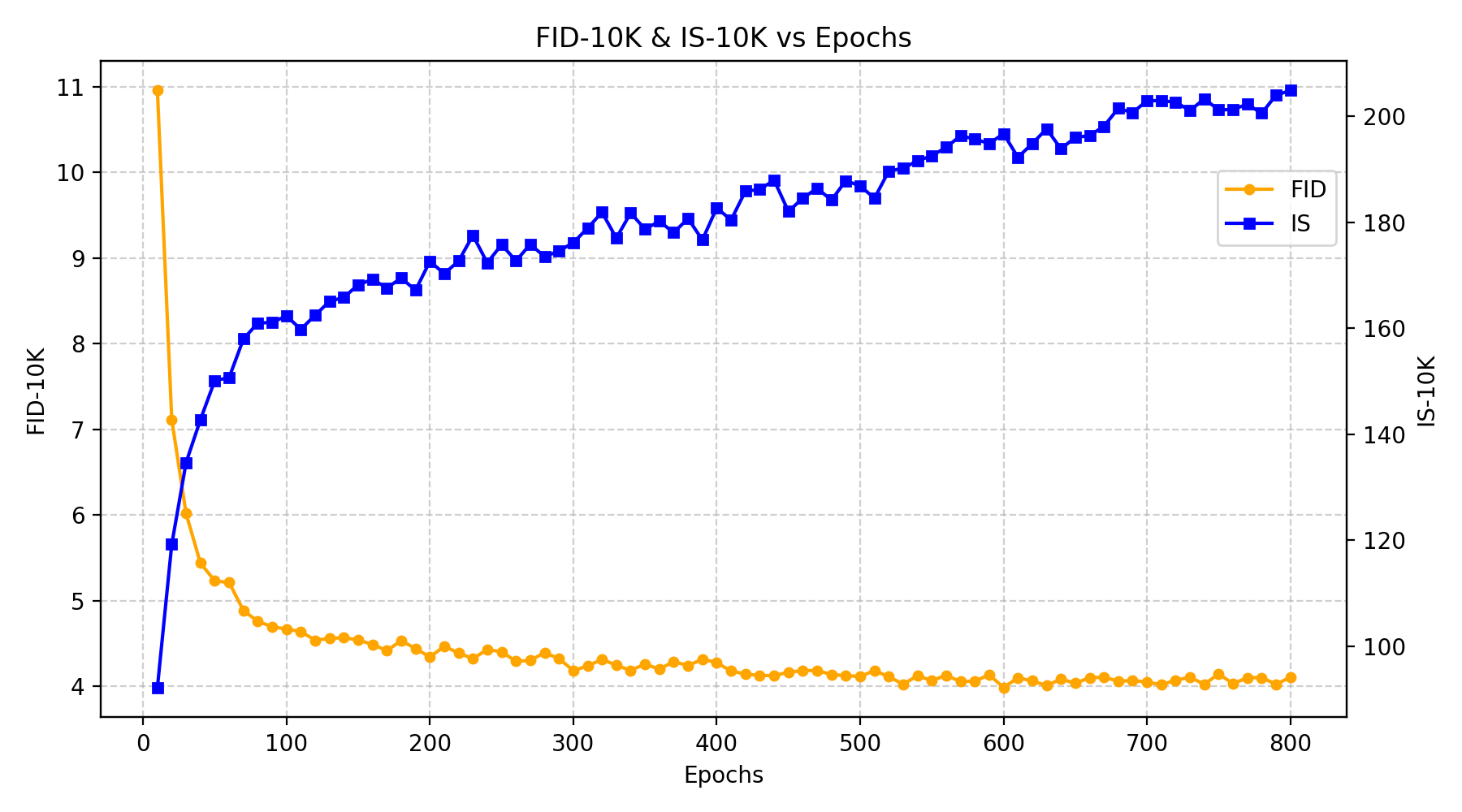}
        \caption{\textbf{FID-10K and IS-10K vs Training Epochs.} FID converges around 600 epochs (between 4.0--4.1), while IS keeps increasing.}
        \label{fig:fid_is_epoch}
    \end{subfigure}\hfill
    \begin{subfigure}[t]{0.48\linewidth}
        \centering
        \includegraphics[width=\linewidth]{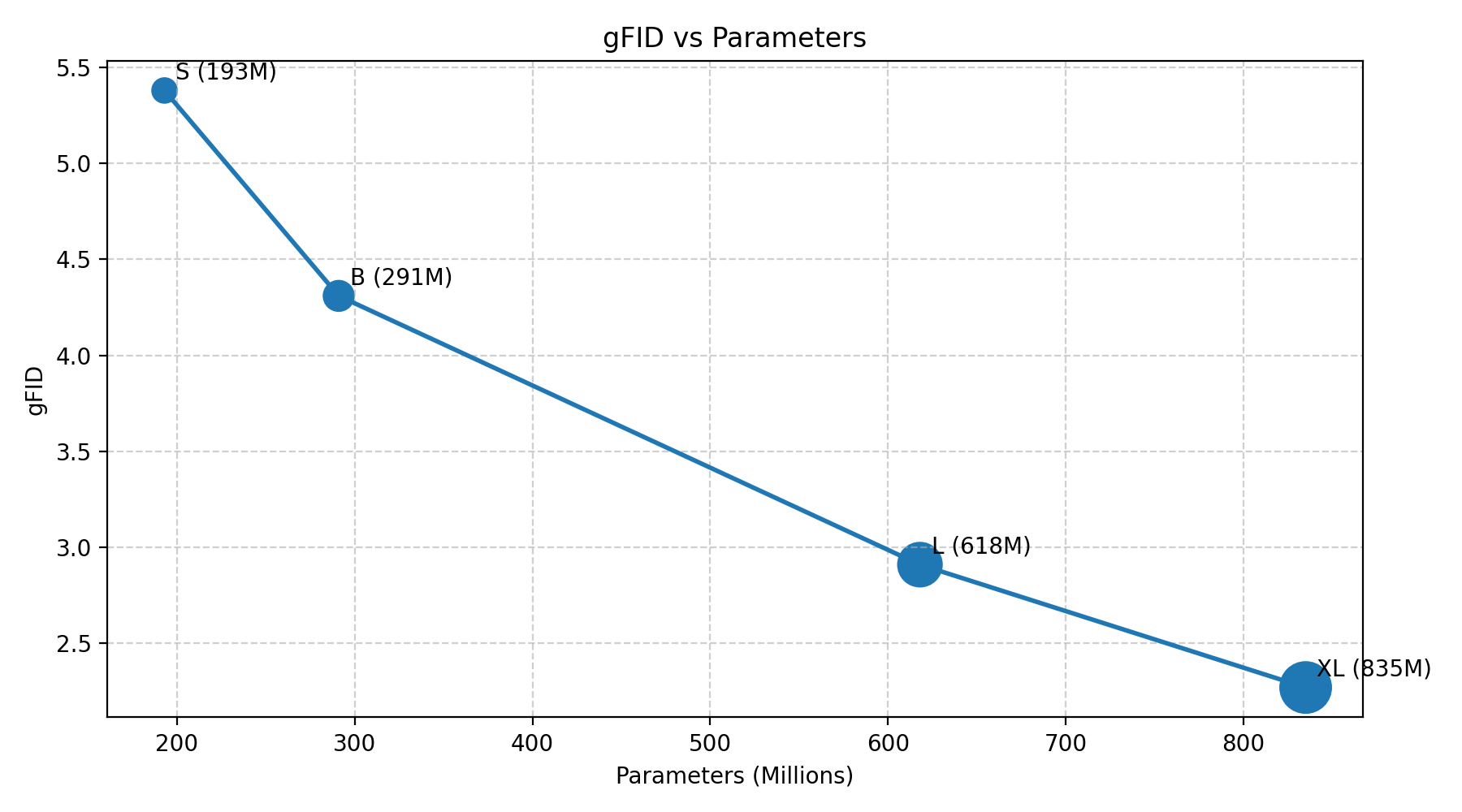}
        \caption{\textbf{FID-50K vs Model Parameters.} Larger models yield lower gFID; evaluated with the 80 epochs checkpoint.}
        \label{fig:fid_params}
    \end{subfigure}

    \vspace{0.6em}

    \begin{subfigure}[t]{0.48\linewidth}
        \centering
        \includegraphics[width=\linewidth]{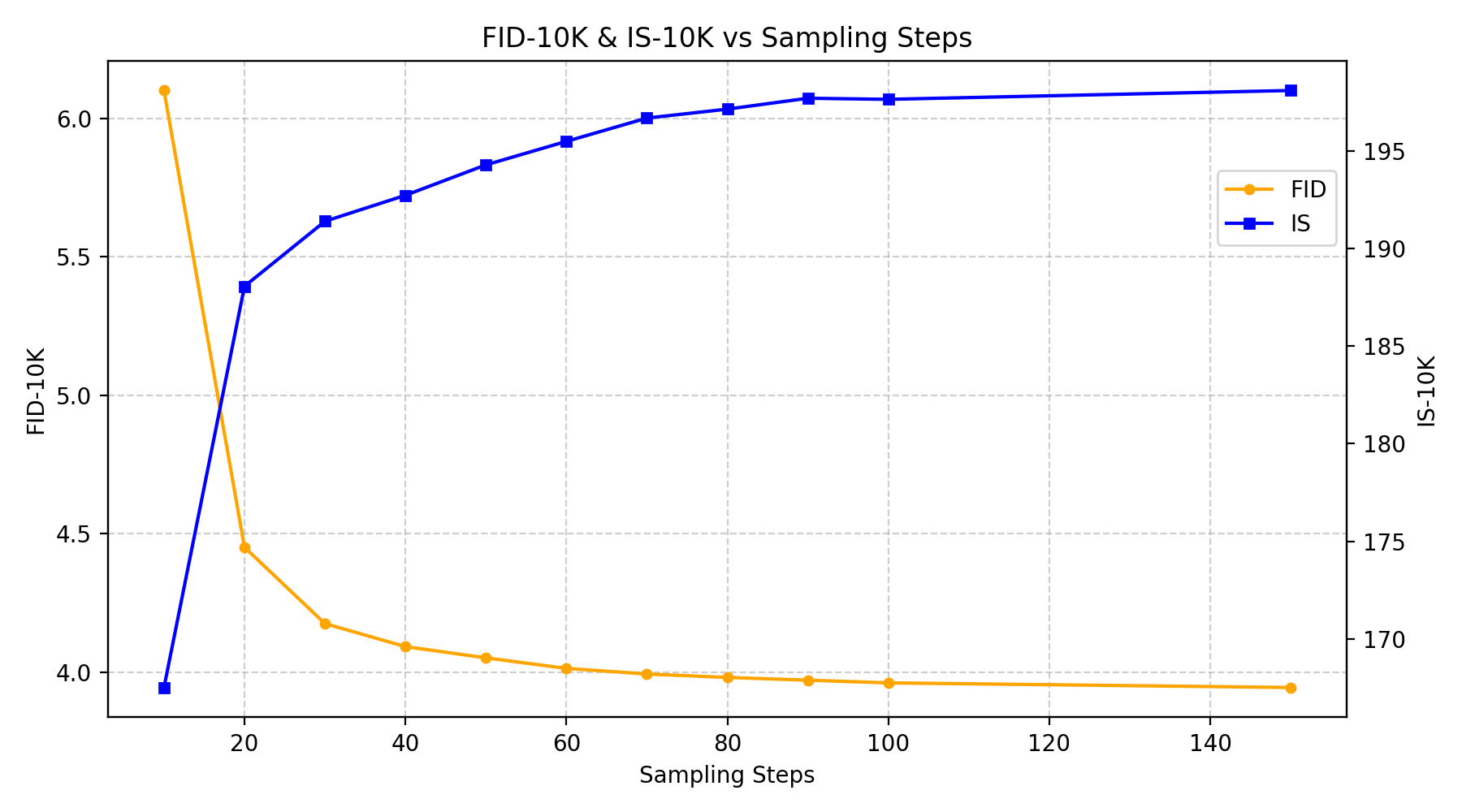}
        \caption{\textbf{FID-10K and IS-10K vs Sampling Steps.} Strong performance is achieved by $\sim$60 steps (trained for 600 epochs).}
        \label{fig:fid_is_steps}
    \end{subfigure}\hfill
    \begin{subfigure}[t]{0.48\linewidth}
        \centering
        \includegraphics[width=\linewidth]{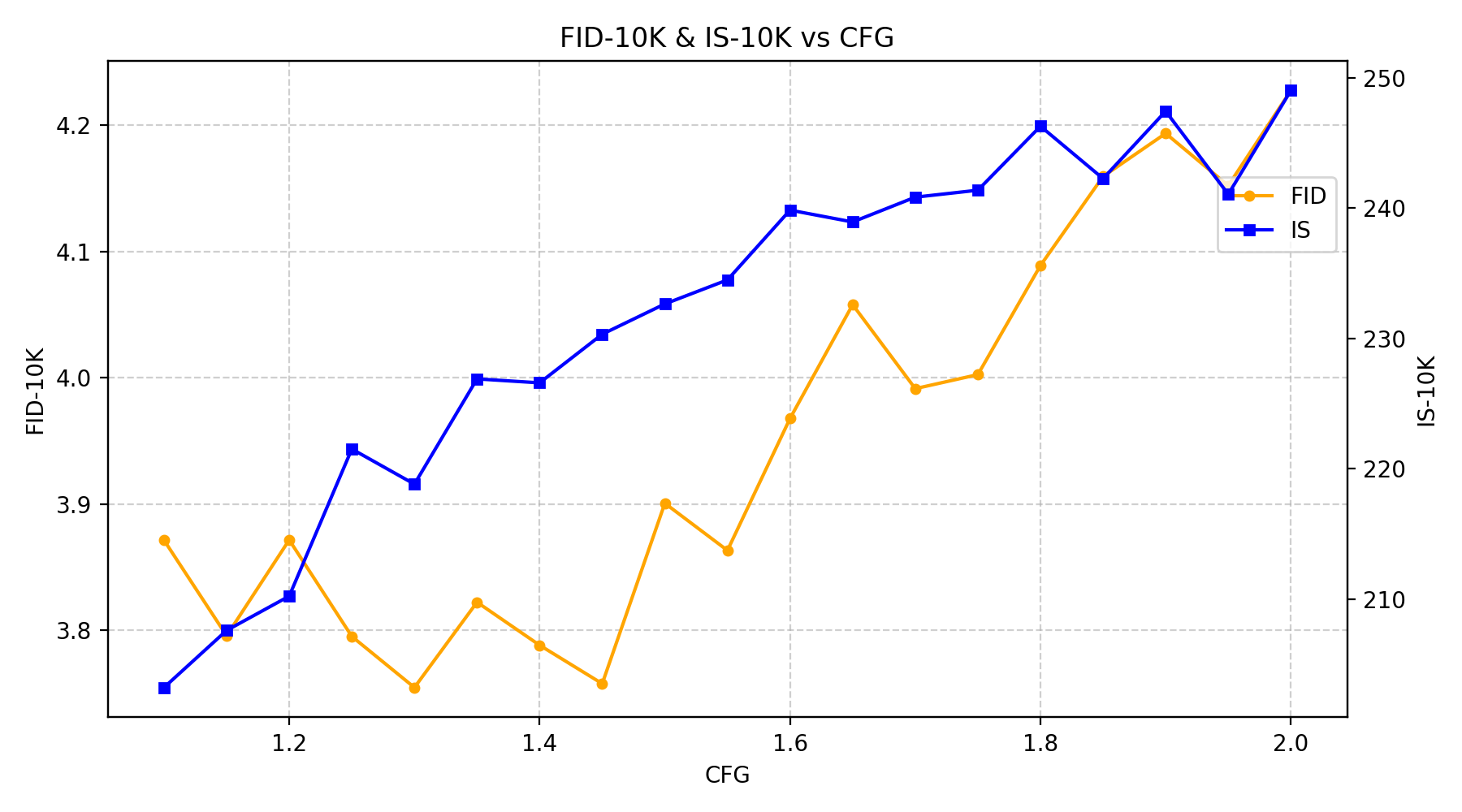}
        \caption{\textbf{FID-10K and IS-10K vs CFG.} Performance across classifier-free guidance scales.}
        \label{fig:fid_is_cfg}
    \end{subfigure}
    \caption{\textbf{FID and IS under different settings.}}
    \vspace{-10pt}
    \label{fig:fid_is_setting}
\end{figure}

\subsection{Main Results}

\noindent
\textbf{Tokenizer Performance.}
In Tab.~\ref{tab:tokenizer_comparison}, we systematically compare RecTok with other representative tokenizers~\cite{LDM, vavae, maetok, RAE, ldetok} in terms of parameter numbers, computational cost, reconstruction, and generation performances.
Although RecTok has more parameters than CNN-based tokenizers, it achieves the lowest computational cost thanks to its efficient ViT architecture. Moreover, ViT-based tokenizers benefit from modern acceleration methods.
In terms of reconstruction, RecTok significantly outperforms other ViT-based methods, while also achieving state-of-the-art performance in generation.
Furthermore, RecTok yields a semantically richer latent space, outperforming prior tokenizers in linear probing (Fig.~\ref{fig:linearprobing}).
In Fig.~\ref{fig:feature_vis}, we visualize the PCA and similarity heatmap, where RecTok exhibits a more semantically structured latent space compared with VA-VAE.
Overall, RecTok achieves the best trade-off among reconstruction fidelity, generation quality, and semantic representation.

In terms of latent dimension, we gradually expand the dimensionality of the latent space.
As shown in Tab.~\ref{tab:dim-results}, we observe a clear trend that the reconstruction, generation, and discriminative performances consistently improve as the dimension increases.
To the best of our knowledge, RecTok is the first work to demonstrate such improvement across all three aspects.

\noindent
\textbf{Generation Comparison.}
As shown in Tab.~\ref{tab:main_res}, RecTok with $\text{DiT}^{\text{DH}}\text{-XL}$ achieves the best gFID=1.34 without classifier-free guidance~\cite{CFG}.
When employing classifier-free guidance, we adopt the AutoGuidance strategy and achieve a gFID of 1.13, matching the gFID of RAE~\cite{RAE} while showing a clear advantage in Inception Score (IS)~\cite{is}.
In Fig.~\ref{fig:fid_is_epoch}, we plot the gFID-10K and IS-10K curves over training epochs.
We observe that gFID-10K stabilizes around 600 epochs, and therefore use the checkpoint at epoch 600 for reporting final results.
The scaling results are presented in Fig.~\ref{fig:fid_params}, where the latent space of RecTok demonstrates strong scaling capability.
Figs.~\ref{fig:fid_is_steps}--\ref{fig:fid_is_cfg} illustrate how different inference settings affect gFID and IS.
Considering the overall performance, we sample 150 steps with a guidance scale of 1.29.

\subsection{Ablation Studies}

\begin{table}[t]
\centering
\caption{
\textbf{Ablations on flow semantic distillation.}
We compare our FSD with the $x_0$ semantic distillation using cosine similarity and VF loss. 
The experimental results demonstrate that FSD yields a significant improvement.
}
\resizebox{1.0\linewidth}{!}{
\begin{tabular}{lccccc}
\toprule
\textbf{Setting} & \textbf{Sem Loss} & \textbf{L.P. Acc.} & \textbf{rFID} & \textbf{gFID} & \textbf{IS} \\
\midrule
\multirow{2}{*}{\makecell[l]{w/o FSD}}
  & Cos Sim & 44.35 & 0.69 & 3.35 & 157.3 \\
  & VF Loss~\cite{vavae} & 37.52 & 0.72 & 3.91 & 142.1 \\
\midrule
\rowcolor{black!6}
w FSD & Cos Sim & \textbf{55.40} & \textbf{0.65} & \textbf{2.27} & \textbf{196.4} \\
\bottomrule
\end{tabular}
}
\label{tab:ablation_path_sem}
\end{table}

\noindent
\textbf{Ablations on FSD.}
In Tab.~\ref{tab:ablation_path_sem}, we study the effectiveness of FSD (i.e., applying distillation on \(x_t\)). When we only align the latent features \(x_0\) to VFM features, a notable degradation of performance in generation and linear probing is observed. This suggests that distilling semantic information along the flow matching path benefits both generation performance and semantic representation. In Tab~\ref{tab:semweightabl}, we compare different $\lambda_{sem}$, considering overall performance, we set $\lambda_{sem}=1$.

\noindent
\textbf{Ablations on Noise Schedule.}
As shown in Tab.~\ref{tab:noiseschedule}, we compare different noise sampling strategies during RecTok training, including the Dimension-dependent Shift (referred to as Shift), Uniform, and Logit-Normal (Lognorm) methods. 
We observe that uniform sampling achieves the best reconstruction performance but performs the worst in generation quality.
In contrast, the shift strategy yields slightly lower reconstruction scores but delivers the best generative results. 
Considering that the reconstruction quality can be further improved through decoder finetuning, as shown in Tab.~\ref{tab:overall_ablation}, we adopt the Shift noise schedule as our default configuration.

\begin{table}[t]
\centering
\caption{\textbf{Ablations on semantic loss weight $\lambda_{sem}$.} $\lambda_{sem}=1$ achieves the best generation performance.}
\footnotesize
\resizebox{0.9\linewidth}{!}{
\begin{tabular}{lcccc}
\toprule
$\lambda_{sem}$ & \textbf{L.P. Acc.} & \textbf{rFID} & \textbf{gFID} & \textbf{IS} \\ 
\midrule
0.5 & 54.8 & \textbf{0.59} & 2.78 & 179.5 \\
\rowcolor{black!6}
1   & 55.4 & 0.65 & \textbf{2.27} & 196.4 \\ 
2   & \textbf{56.1} & 0.87 & 2.43 & \textbf{199.7} \\
\bottomrule
\end{tabular}
}
\label{tab:semweightabl}
\end{table}

\begin{table}[t]
\centering
\caption{\textbf{Ablations on vision foundation models (VFMs).} 
DINOv2 excels in low-dimensional latents (e.g., 16), while DINOv3 performs best in higher dimensions (e.g., 128).}
\setlength{\tabcolsep}{8pt}
\footnotesize
\resizebox{1.0\linewidth}{!}{
\begin{tabular}{lcccccc}
\toprule
\multirow{2}{*}{\textbf{VFM}} & \multicolumn{3}{c}{\textbf{Dim=16}} & \multicolumn{3}{c}{\textbf{Dim=128}} \\
\cmidrule(lr){2-4}
\cmidrule(lr){5-7}
& \textbf{rFID} & \textbf{gFID} & \textbf{IS} & \textbf{rFID} & \textbf{gFID} & \textbf{IS} \\
\midrule
\rowcolor{black!6}
DINOv3~\cite{dinov3}         & 0.81 & 2.86          & \textbf{195.2} & 0.65 & \textbf{2.27} & \textbf{196.4} \\
DINOv2~\cite{Dinov2}         & 0.74 & \textbf{2.75} & 183.3 & 0.53 & 2.38          & 184.7 \\
SigLIP 2~\cite{siglip2}      & 0.71 & 3.59          & 172.3 & 0.51 & 3.14          & 178.6 \\
RADIOv2~\cite{radio}         & 0.79 & 2.97          & 193.1 & 0.64 & 2.59          & 193.4 \\
SAM~\cite{sam}               & 0.83 & 4.96          & 141.1 & 0.69 & 4.47          & 157.2 \\
\midrule
Two VFMs                     & \textbf{0.69} & 3.26          & 164.7 & \textbf{0.49} & 2.51          & 181.3 \\
\bottomrule
\end{tabular}
}
\label{tab:vfm_ablation}
\footnotesize{$^\dagger$Using two VFMs simultaneously (DINOv3 and SigLIP 2)}
\end{table}

\noindent
\textbf{Ablations on VFMs.}
We explore the impact of different VFMs in FSD, including DINOv3~\cite{dinov3}, DINOv2~\cite{Dinov2}, SigLIP2~\cite{siglip2}, RADIOv2~\cite{radio}, and SAM~\cite{sam}. All models use the large-size variants to align with the VA-VAE experiments. We find that DINOv3 achieves the best performance in high-dimensional latent spaces, while DINOv2 performs best in low-dimensional latent spaces. In particular, we also try using two VFMs during semantic learning. However, this leads to a degradation in generation performance. Considering these results, we use DINOv2 for experiments with 16 and 32 dimensions, and DINOv3 for experiments with 64 dimensions and above.

\begin{table}[t]
\centering
\caption{\textbf{Ablations on reconstruction and alignment distillation.}
RAD improves the generation performance, and the gain does not originate from the transformer architecture.
}
\label{tab:ablation_align_recon}
\setlength{\tabcolsep}{8pt}
\resizebox{1.0\linewidth}{!}{
\begin{tabular}{lcccc}
\toprule
\textbf{Setting} & \textbf{Sem Dec} & \textbf{rFID} & \textbf{gFID} & \textbf{IS} \\
\midrule
Align. only & MLP         & 0.76 & 3.02 & 175.3 \\
Align. only & Transformer & \textbf{0.57} & 2.52 & 184.5 \\
Rec. only & Transformer        & 0.75 & 2.97 & 174.2 \\
\rowcolor{black!6}
Rec. + Align. & Transformer   & 0.65 & \textbf{2.27} & \textbf{196.4} \\
\bottomrule
\end{tabular}
}
\end{table}

\noindent
\textbf{Ablations on RAD.}
We conduct four groups of ablation studies, including (1) alignment only, (2) reconstruction only, and (3) joint reconstruction and alignment. For the alignment only setting, we experiment with two types of semantic decoders: an MLP (4M parameters) and a lightweight Transformer (1.5M parameters). As shown in Table~\ref{tab:ablation_align_recon}, the joint reconstruction and alignment strategy achieves the best overall performance, obtaining the lowest gFID, and the highest IS score.

\begin{table}[t]
\centering
\caption{\textbf{Comparison of different sampling distributions.}
Since the reconstruction quality can be improved through decoder finetuning, we adopt the Shift schedule.
}
\resizebox{1.0\linewidth}{!}{
\begin{tabular}{lccccc}
\toprule
\textbf{Noise Schedule} & \textbf{L.P. Acc.} & \textbf{rFID} & \textbf{PSNR} & \textbf{gFID} & \textbf{IS} \\
\midrule
Uniform & 55.1 & \textbf{0.53} & \textbf{26.71} & 2.50 & 191.6 \\
Lognorm & 53.5 & 0.57 & 25.03 & 2.37 & \textbf{199.7} \\
\rowcolor{black!6}
Shift   & \textbf{55.4} & 0.65 & 25.28 & \textbf{2.27} & 196.4 \\
\bottomrule
\end{tabular}
}
\label{tab:noiseschedule}
\end{table}

\begin{table}[t]
    \centering
    \caption{\textbf{Ablations on encoder initialization methods.}
    We notice that using the randomly initialized encoder yields better generation performance.
    }
    \resizebox{1.0\linewidth}{!}{
    \begin{tabular}{lcccc}
    \toprule
    \textbf{Encoder Initialization} & \textbf{L.P. Acc.} & \textbf{rFID} & \textbf{gFID} & \textbf{IS} \\
    \midrule
    VFM    & 54.5 & \textbf{0.57} & 2.37 & 189.2 \\
    \rowcolor{black!6}
    Random & \textbf{55.4} & 0.65 & \textbf{2.27} & \textbf{196.4} \\
    \bottomrule
    \end{tabular}
    }
    \label{tab:encoder_ablation_table}
\end{table}

\begin{table}[t]
\centering
\caption{\textbf{Comparison of the performance on different semantic decoders.}
A lightweight transformer achieves the best generation performance.
}
\resizebox{1.0\linewidth}{!}{
\begin{tabular}{lccccc}
\toprule
\textbf{Sem Dec} & \textbf{Params} & \textbf{L.P. Acc.} & \textbf{rFID} & \textbf{gFID} & \textbf{IS} \\
\midrule
MLP   & 4M   & 51.2 & 0.76 & 3.02 & 175.3 \\
Transformer & 10M  & 47.3 & \textbf{0.63} & 3.13 & 170.5 \\
\rowcolor{black!6}
Transfromer & 1.5M & \textbf{55.4} & 0.65 & \textbf{2.27} & \textbf{196.4} \\
\bottomrule
\end{tabular}
}
\label{tab:semdec_comparison}
\end{table}

\noindent
\textbf{Ablations on Semantic Decoder.}
In Tab.~\ref{tab:semdec_comparison}, we compare different architectural designs of the Semantic Decoder. 
We observe that using a transformer architecture consistently outperforms an MLP across all metrics. 
However, increasing the transformer's capacity leads to degraded performance in both linear probing accuracy and generation quality.
Therefore, a lightweight transformer design provides the best overall trade-off.

\noindent
\textbf{Ablations on Encoder Initialization.}
We attempt to initialize the encoder of RecTok with VFMs, as shown in Tab.~\ref{tab:encoder_ablation_table}. 
We employ FSD and RAD as self-distillation~\cite{clipself} like methods to train the VFM initialized RecTok. 
However, the overall performance lags behind that of the randomly initialized encoder.


\begin{table}[t]
\centering
\caption{\textbf{Overall ablation study of FSD, RAD, and Decoder Finetuning.}
Each method provides a clear improvement.
}
\resizebox{1.0\linewidth}{!}{
\begin{tabular}{lccccc}
\toprule
\textbf{Method} & \textbf{L.P. Acc.} & \textbf{rFID} & \textbf{PSNR} & \textbf{gFID} & \textbf{IS} \\
\midrule
Baseline & 7.1  & \textbf{0.22} & \textbf{29.76} & 12.07 & 57.5 \\
+ FSD    & 52.7 & 0.57 & 25.62 & 2.52  & 184.5 \\
+ RAD    & \textbf{55.4} & 0.65 & 25.28 & 2.27  & 196.4 \\
\rowcolor{black!6}
+ Dec FT & \textbf{55.4} & 0.48 & 26.16 & \textbf{2.23}  & \textbf{198.2} \\
\bottomrule
\end{tabular}
}
\vspace{-5pt}
\label{tab:overall_ablation}
\end{table}

\noindent
\textbf{Overall Ablation Study.}
In Tab.~\ref{tab:overall_ablation}, we present an ablation study on the two key innovations and the decoder finetuning. 
Each component brings a clear performance gain.
\section{Conclusion}
\label{sec:conclusion}
 
In this work, we address the fundamental challenge posed by the latent dimensionality of visual tokenizers through RecTok.
Building on our core insight---enhancing semantic consistency along the forward flow rather than only at the un-noised latents. We introduce two key innovations: Flow Semantic Distillation (FSD) and Reconstruction and Alignment Distillation (RAD).
Together, FSD and RAD effectively enrich the semantics of RecTok's latent space, and we observe consistent improvements as the latent dimension increases.
Experiments on ImageNet-1K demonstrate that RecTok achieves state-of-the-art generation performance while maintaining strong reconstruction quality and semantic representation. 
We hope this work inspires future research on high-dimensional visual tokenizers.

{
    \small
    \bibliographystyle{ieeenat_fullname}
    \bibliography{main}
}

\clearpage
\setcounter{page}{1}
\maketitlesupplementary

\noindent
\textbf{Overview.} In this supplementary file, we present more details in addition to the main paper. Here are the details:
\begin{itemize}
    \item \textbf{\cref{sec:introduction_video}}: Introduction video, we strongly recommend that reviewers take a look.
    \item \textbf{\cref{sec:impl_details}}: Implementation details.
    \item \textbf{\cref{sec:additional_ablation_studies}}: More ablation studies.
    \item \textbf{\cref{sec:additional_qualitative_results}}: More qualitative results.
    \item \textbf{\cref{sec:additional_analysis_rectok}}: More analysis on RecTok.
    \item \textbf{\cref{sec:limitations}}: Limitations and future works.
    \item \textbf{\cref{sec:broader_impacts}}: Broader impacts.
\end{itemize}

\section{Introduction Video}
\label{sec:introduction_video}
To help readers quickly grasp the primary idea of our work, we provide a 6-minute introduction video. Please refer to ``\textbf{introduction\_video.mp4}'' in the supplementary file.

\section{Implementation Details}
\label{sec:impl_details}
In the Tab.~\ref{tab:impl_details}, we provide detailed training configurations for both RecTok and DiT, including the hyperparameters for different DiT sizes, training epochs, learning-rate schedules, sampling schedules, and other related settings.

\section{Additional Ablation Studies}
\label{sec:additional_ablation_studies}

\noindent
\textbf{Ablations on Mask Ratio in RAD.}
We ablate the mask ratio used in reconstruction and alignment distillation (RAD). As shown in Tab.~\ref{tab:mask_ratio_abl}, we evaluate four mask ratio settings and report their effects on reconstruction, generation, and semantics. 
The results show that increasing the mask ratio improves generation quality while degrading reconstruction performance. Since reconstruction can be further enhanced through decoder finetuning, we set the upper bound of the mask ratio to 0.4.

\noindent
\textbf{Ablations on Noise Intensity in FSD.}
It is worth noting that the end point state of the forward flow in FSD, $x_1$, does not need to follow a normal Gaussian distribution. As an alternative, we can instead place $x_1$ in a more expressive high-intensity noise space:
\begin{equation}
    x_1 = \gamma \times \epsilon, \quad \epsilon \in \mathcal{N}(0, 1),
\end{equation}

where $\epsilon$ denotes standard Gaussian noise, and $\gamma$ controls the intensity of the noise, equivalently, the variance of $x_1$. A larger $\gamma$ corresponds to stronger collisions at time $t$. 
We view this mechanism as analogous to the timestep shift introduced in our main paper, as both aim to mitigate information redundancy in high-dimensional settings. Therefore, we focus our analysis on the effect of $\gamma$ and leave alternative formulations of $x_1$ to future work.

\noindent
\textbf{Ablations on KL Loss.}
We conduct an ablation study on the use of the KL loss as shown in Tab.~\ref{tab:kl_loss_abl}. Although recent works~\cite{dinov2l, maetok} remove the KL term and adopt an autoencoder (AE), our ablations show that incorporating KL regularization improves the generation performance.
A detailed analysis of Tab.~\ref{tab:kl_loss_abl} reveals a distinct trade-off between reconstruction and generation performance. Specifically, the deterministic AE setting excels in reconstruction, achieving a lower rFID (0.35) and higher PSNR (29.89). This indicates that without the regularization constraint, the model can more freely encode high-frequency details into the latent space. However, this unconstrained latent space fall short in the generation stage, as evidenced by the degraded gFID (5.19). In contrast, enforcing the KL loss promotes a smooth and compact latent manifold. Although this results in a slight drop in reconstruction metrics, it significantly facilitates the learning process for the subsequent generative model, improving the gFID by over 50\% (5.19 $\rightarrow$ 2.27) and boosting the Inception Score by a large margin (+44.2).
Since our work prioritizes the generation task, we retain the KL loss and use a variational autoencoder (VAE).

\begin{table}[t]
\centering
\small
\caption{\textbf{Implementation details.} We report the detailed architectural specifications, training hyperparameters, and sampling settings for different model variants.}
\setlength{\tabcolsep}{6pt}
\resizebox{1.0\linewidth}{!}{
\begin{tabular}{lcccc}
\toprule
\textbf{architecture} & \textbf{$\text{DiT}^{\text{DH}}$-S} & \textbf{$\text{DiT}^{\text{DH}}$-B} & \textbf{$\text{DiT}^{\text{DH}}$-L} & \textbf{$\text{DiT}^{\text{DH}}$-XL} \\
\midrule
depth & 12 & 12 & 24 & 28 \\
hidden dim & 384 & 768 & 1024 & 1152 \\
heads & 6 & 12 & 16 & 16 \\
DDT depth & \multicolumn{4}{c}{2} \\
DDT hidden dim & \multicolumn{4}{c}{2048} \\
DDT heads & \multicolumn{4}{c}{16} \\
\midrule

\textbf{training} & \multicolumn{2}{c}{RecTok} & \multicolumn{2}{c}{$\text{DiT}^{\text{DH}}$} \\
\midrule
epochs & \multicolumn{2}{c}{200} & \multicolumn{2}{c}{80 (ablation), 600} \\
warmup epochs & \multicolumn{2}{c}{50 (linear)} & \multicolumn{2}{c}{0} \\
decay epochs & \multicolumn{2}{c}{50 - 200} & \multicolumn{2}{c}{40 - 800} \\
optimizer & \multicolumn{4}{c}{Adam~\cite{adam}, $\beta_{1}, \beta_{2} = 0.9, 0.95$} \\
batch size & \multicolumn{4}{c}{1024} \\
learning rate & \multicolumn{2}{c}{4e-4} & \multicolumn{2}{c}{2e-4} \\
learning rate schedule & \multicolumn{2}{c}{cosine decay} & \multicolumn{2}{c}{linear decay} \\
weight decay & \multicolumn{2}{c}{1e-4} & \multicolumn{2}{c}{0.0} \\
ema rate & \multicolumn{2}{c}{0.999} & \multicolumn{2}{c}{0.9995} \\
noise schedule & \multicolumn{4}{c}{$t = \frac{s t'}{1 + (s - 1)t'}, \quad t' \sim \mathcal{U}(0,1), \quad s = \sqrt{\frac{4096}{r^2 d}}$} \\
class token drop (for CFG) & \multicolumn{2}{c}{None} & \multicolumn{2}{c}{0.1} \\
\midrule

\multicolumn{5}{l}{\textbf{sampling}} \\
\midrule
ODE solver & \multicolumn{4}{c}{Euler} \\
ODE steps & \multicolumn{4}{c}{50 (ablation), 150} \\
time steps & \multicolumn{4}{c}{shift in [0.0, 1.0] according to noise schedule} \\
CFG scale  & \multicolumn{4}{c}{1.29} \\
CFG interval & \multicolumn{4}{c}{[0, 1] (not used)} \\
\bottomrule
\end{tabular}
}
\label{tab:impl_details}
\end{table}

\begin{table}[t]
\centering
\caption{\textbf{Ablation on mask ratio.} We observe that a higher mask ratio of 0.4 yields the best overall generative performance (gFID and IS), suggesting that a more challenging task benefits RAD.}
\small
\resizebox{0.9\linewidth}{!}{
\begin{tabular}{lcccc}
\toprule
\textbf{Mask Ratio} & \textbf{rFID} & \textbf{PSNR} & \textbf{gFID} & \textbf{IS} \\
\midrule
0.3 & \textbf{0.60} & \textbf{25.45} & 2.35 & 192.8 \\
\rowcolor{black!6}
0.4 & 0.65 & 25.28 & \textbf{2.27} & 196.4 \\
0.5 & 0.66 & 25.24 & 2.28 & \textbf{197.1} \\
0.6 & 0.67 & 25.22 & 2.27 & 192.7 \\
\bottomrule
\end{tabular}
}
\label{tab:mask_ratio_abl}
\end{table}


\begin{table}[t]
\centering
\small
\caption{\textbf{Effect of $\gamma$ on reconstruction and generation quality.} We observe that $\gamma=1.0$ achieves the optimal trade-off. While larger $\gamma$ values marginally improve IS, they lead to a degradation in both reconstruction fidelity (rFID/PSNR) and generative distribution alignment (gFID).}
\setlength{\tabcolsep}{6pt}
\resizebox{0.9\linewidth}{!}{
\begin{tabular}{lccccc}
\toprule
\textbf{$\gamma$} & \textbf{Noise Schedule} & \textbf{rFID} & \textbf{PSNR} & \textbf{gFID} & \textbf{IS} \\
\midrule
\rowcolor{black!6}
1.0 & Shift & \textbf{0.65} & \textbf{25.28} & \textbf{2.27} & 196.4 \\
2.0 & Shift & 0.69 & 24.79 & 2.34 & 198.1 \\
3.0 & Shift & 0.72 & 24.45 & 2.39 & \textbf{200.2} \\
\bottomrule
\end{tabular}
}
\end{table}

\begin{table}[t]
\centering
\caption{\textbf{Ablation on the KL loss.} We compare the deterministic AE (w/o KL) and the VAE (w KL) settings. While removing the KL term improves reconstruction fidelity (rFID 0.35), it results in a disjointed latent space that hinders generation (gFID 5.19). Incorporating KL regularization significantly boosts generative performance (gFID 2.27), validating its necessity for our framework.}
\small
\resizebox{0.9\linewidth}{!}{
\begin{tabular}{lcccc}
\toprule
\textbf{Setting} & \textbf{rFID} & \textbf{PSNR} & \textbf{gFID} & \textbf{IS} \\
\midrule
w/o KL & \textbf{0.35} & \textbf{29.89} & 5.19 & 152.2 \\
\rowcolor{black!6}
w KL   & 0.65 & 25.28 & \textbf{2.27} & \textbf{196.4} \\
\bottomrule
\end{tabular}
}
\label{tab:kl_loss_abl}
\end{table}

\begin{figure*}[t]
    \centering
    
    \begin{subfigure}[b]{0.19\textwidth}
        \centering
        \includegraphics[width=\textwidth]{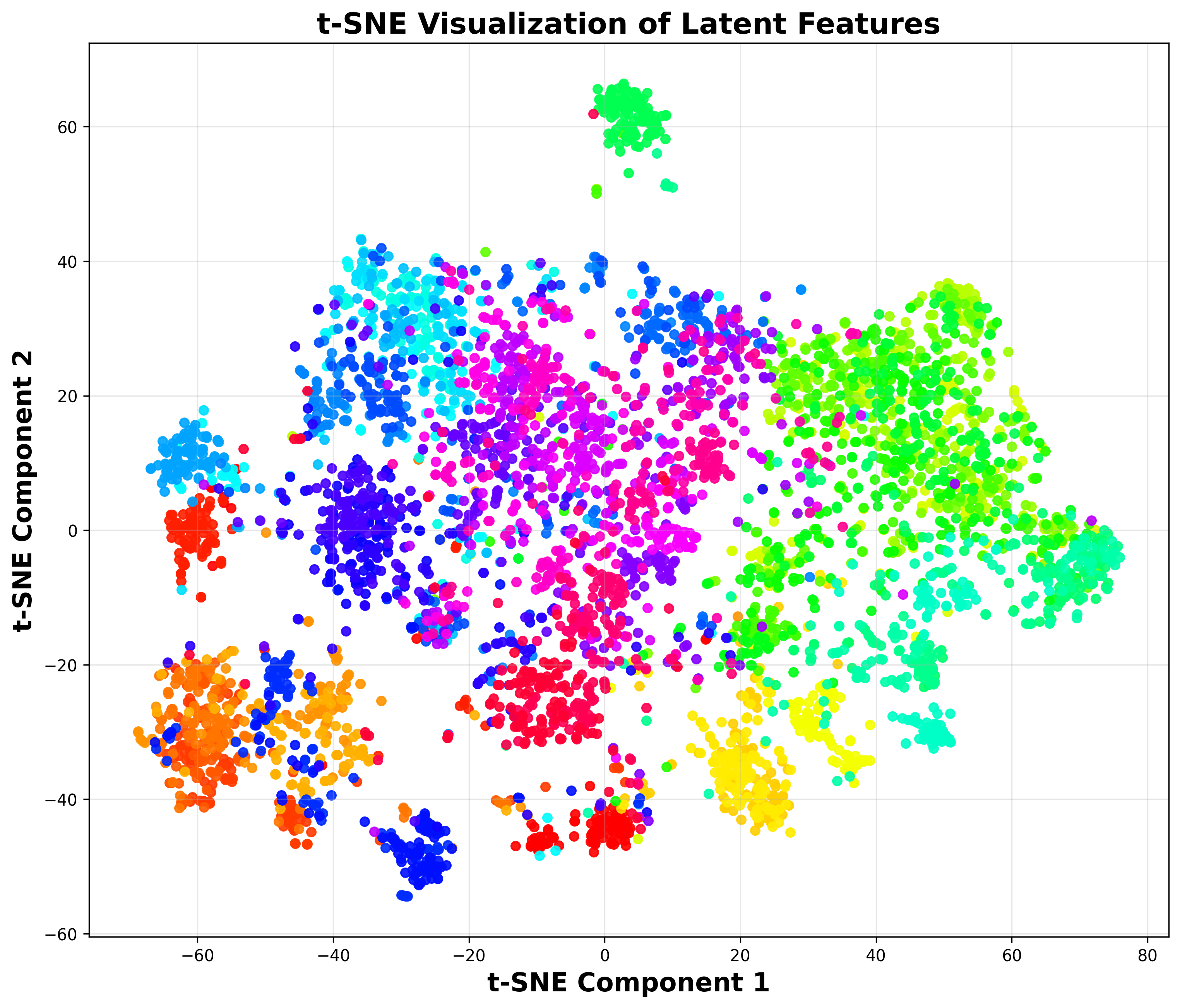}
        \caption{t=0.2, acc.=52.8\%}
    \end{subfigure}
    \begin{subfigure}[b]{0.19\textwidth}
        \centering
        \includegraphics[width=\textwidth]{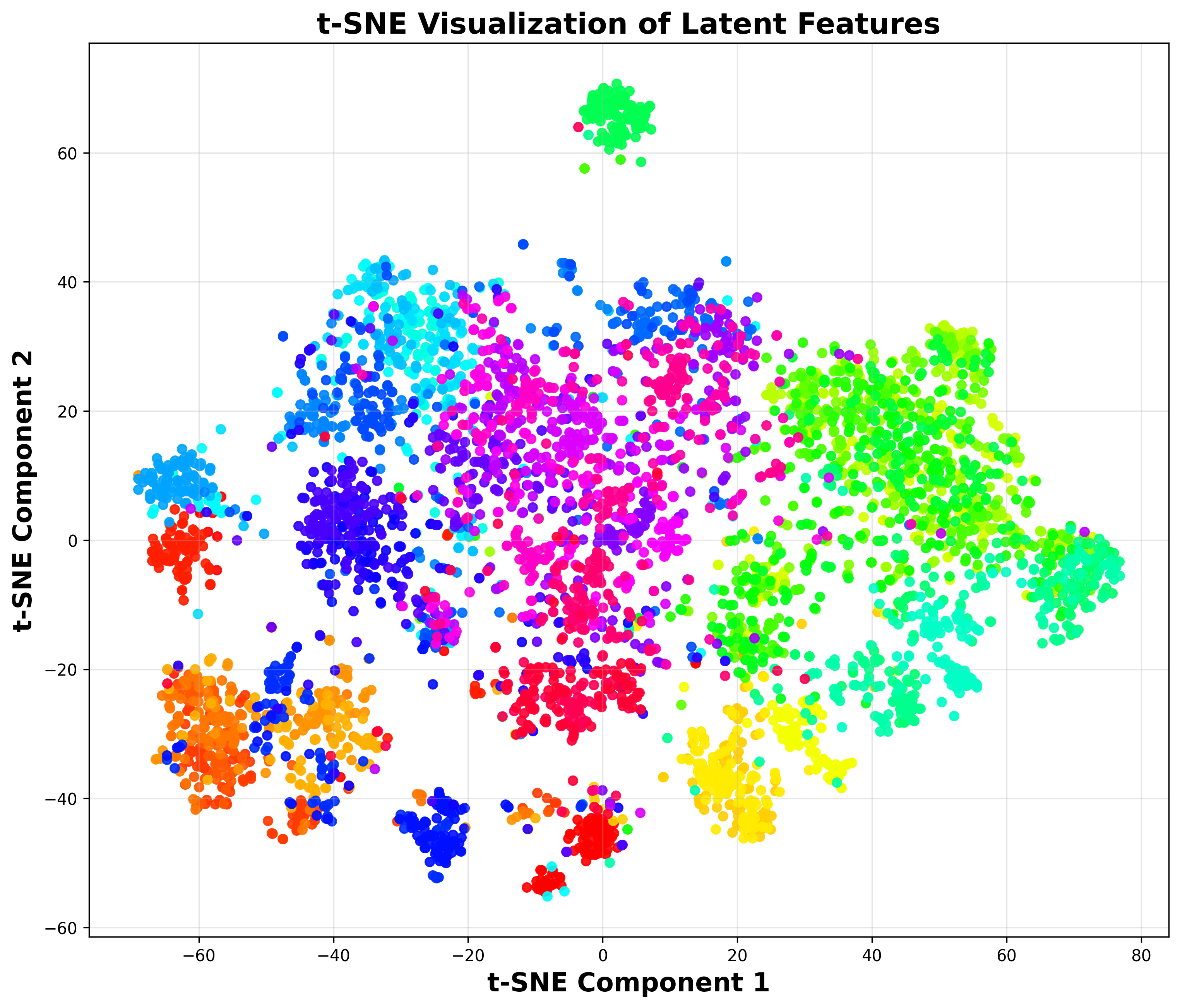}
        \caption{t=0.4, acc.=54.9\%}
    \end{subfigure}
    \begin{subfigure}[b]{0.19\textwidth}
        \centering
        \includegraphics[width=\textwidth]{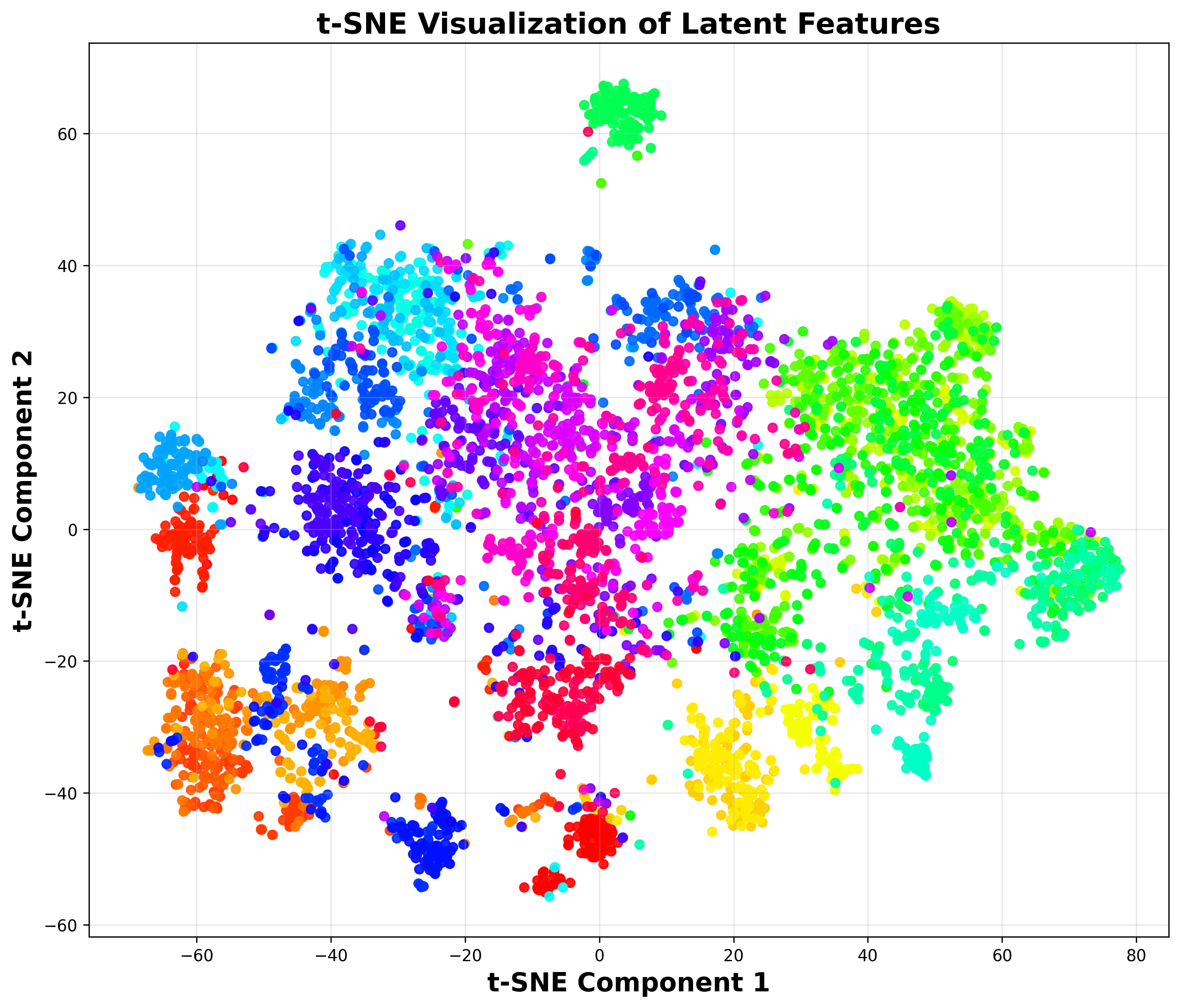}
        \caption{t=0.6, acc.=55.5\%}
    \end{subfigure}
    \begin{subfigure}[b]{0.19\textwidth}
        \centering
        \includegraphics[width=\textwidth]{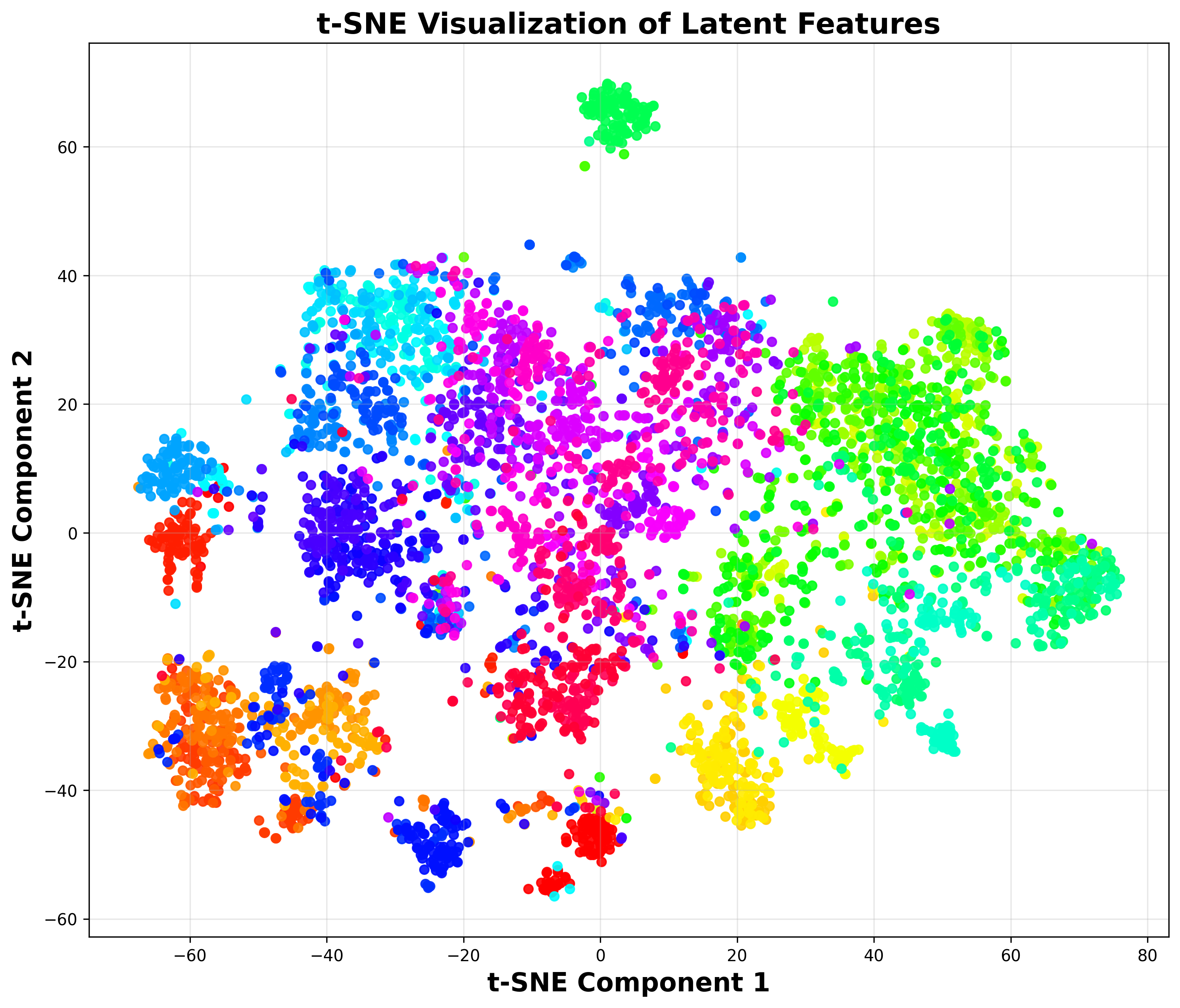}
        \caption{t=0.8, acc.=55.2\%}
    \end{subfigure}
    \begin{subfigure}[b]{0.19\textwidth}
        \centering
        \includegraphics[width=\textwidth]{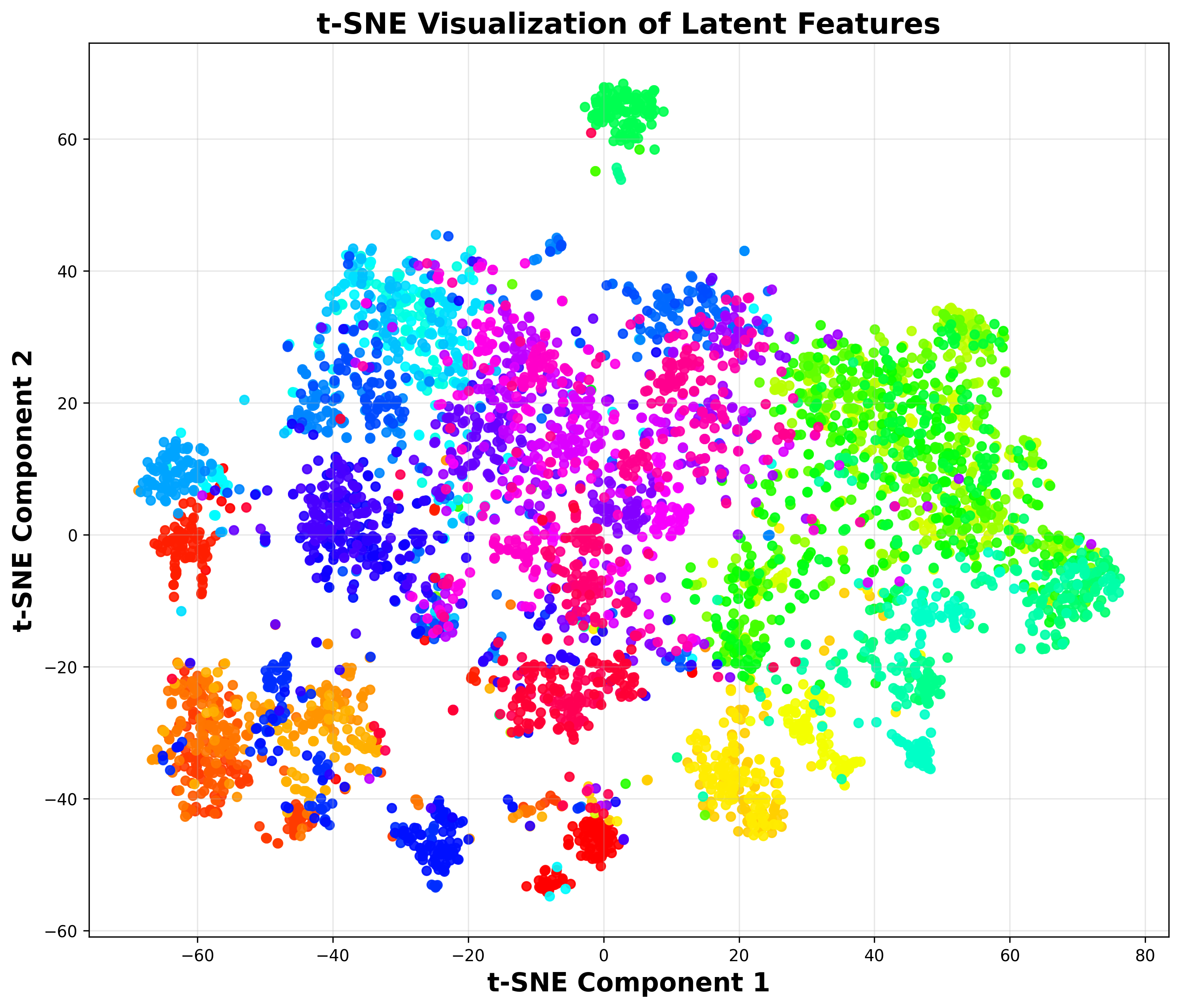}
        \caption{t=1.0, acc.=55.4\%}
    \end{subfigure}
    \caption{\textbf{t-SNE visualizations under different timesteps (t).} Our RecTok shows a clear advantage in semantic consistency on the forward flow, even with a high level of noise and disturbance.}
    \label{fig:flow}
\end{figure*}

\begin{figure*}[t]
    \centering
    
    \begin{minipage}[c]{0.02\textwidth}
        \centering
        \rotatebox{90}{\textbf{RecTok}}
    \end{minipage}
    \hfill
    \begin{minipage}[c]{0.97\textwidth}
        \centering
        \begin{subfigure}[b]{0.19\textwidth}
            \centering
            \includegraphics[width=\textwidth]{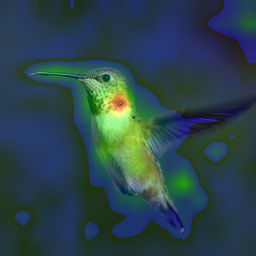}
        \end{subfigure}
        \hfill
        \begin{subfigure}[b]{0.19\textwidth}
            \centering
            \includegraphics[width=\textwidth]{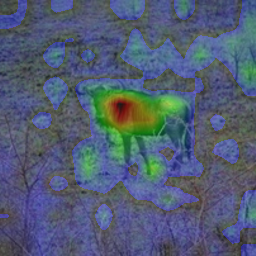}
        \end{subfigure}
        \hfill
        \begin{subfigure}[b]{0.19\textwidth}
            \centering
            \includegraphics[width=\textwidth]{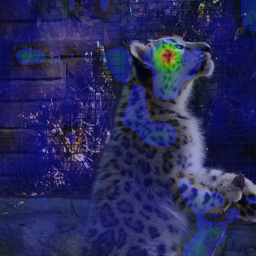}
        \end{subfigure}
        \hfill
        \begin{subfigure}[b]{0.19\textwidth}
            \centering
            \includegraphics[width=\textwidth]{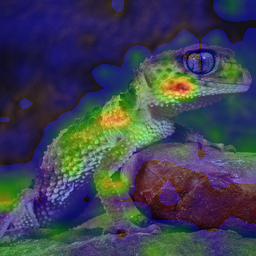}
        \end{subfigure}
        \hfill
        \begin{subfigure}[b]{0.19\textwidth}
            \centering
            \includegraphics[width=\textwidth]{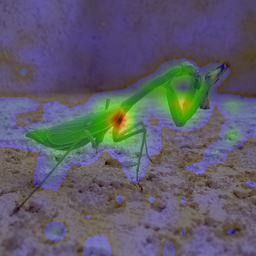}
        \end{subfigure}
    \end{minipage}

    \vspace{2pt} 

    \begin{minipage}[c]{0.02\textwidth}
        \centering
        \rotatebox{90}{\textbf{VA-VAE}~\cite{vavae}}
    \end{minipage}
    \hfill
    \begin{minipage}[c]{0.97\textwidth}
        \centering
        \begin{subfigure}[b]{0.19\textwidth}
            \centering
            \includegraphics[width=\textwidth]{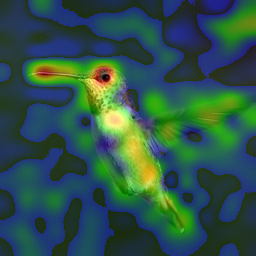}
        \end{subfigure}
        \hfill
        \begin{subfigure}[b]{0.19\textwidth}
            \centering
            \includegraphics[width=\textwidth]{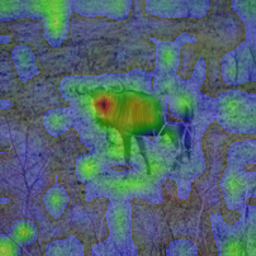}
        \end{subfigure}
        \hfill
        \begin{subfigure}[b]{0.19\textwidth}
            \centering
            \includegraphics[width=\textwidth]{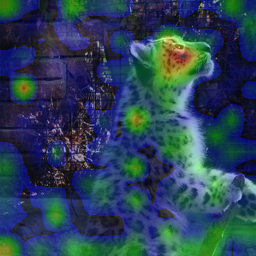}
        \end{subfigure}
        \hfill
        \begin{subfigure}[b]{0.19\textwidth}
            \centering
            \includegraphics[width=\textwidth]{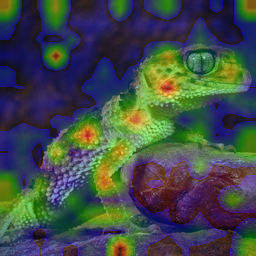}
        \end{subfigure}
        \hfill
        \begin{subfigure}[b]{0.19\textwidth}
            \centering
            \includegraphics[width=\textwidth]{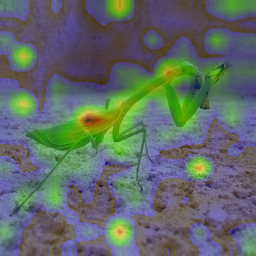}
        \end{subfigure}
    \end{minipage}
    
    \caption{\textbf{Visualizations of latent feature through cosine similarity.} We compare the latent features of RecTok (top row) and VA-VAE (bottom row). The RecTok features exhibit stronger semantic localization on foreground objects compared to VA-VAE.}
    \label{fig:feat_cos}
\end{figure*}

\begin{figure*}[t]
    \centering
    
    \begin{subfigure}[b]{0.19\textwidth}
        \centering
        \includegraphics[width=\textwidth]{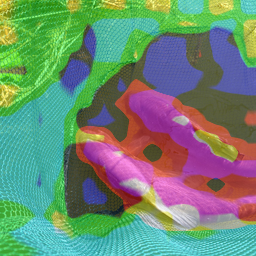}
    \end{subfigure}
    \begin{subfigure}[b]{0.19\textwidth}
        \centering
        \includegraphics[width=\textwidth]{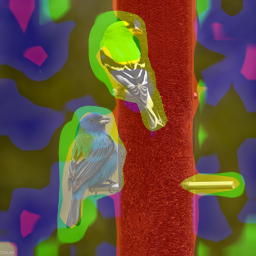}
    \end{subfigure}
    \begin{subfigure}[b]{0.19\textwidth}
        \centering
        \includegraphics[width=\textwidth]{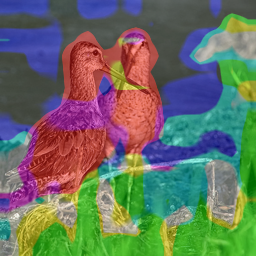}
    \end{subfigure}
    \begin{subfigure}[b]{0.19\textwidth}
        \centering
        \includegraphics[width=\textwidth]{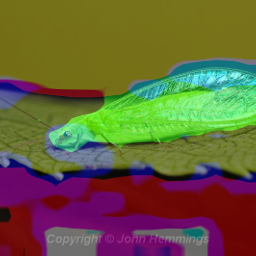}
    \end{subfigure}
    \begin{subfigure}[b]{0.19\textwidth}
        \centering
        \includegraphics[width=\textwidth]{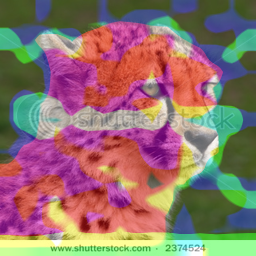}
    \end{subfigure}
    \caption{\textbf{Visualizations of latent feature through PCA.}}
    \label{fig:feat_pca}
\end{figure*}

\section{Additional Qualitative Results}
\label{sec:additional_qualitative_results}

\begin{figure*}[p]
\centering

\includegraphics[width=0.95\linewidth]{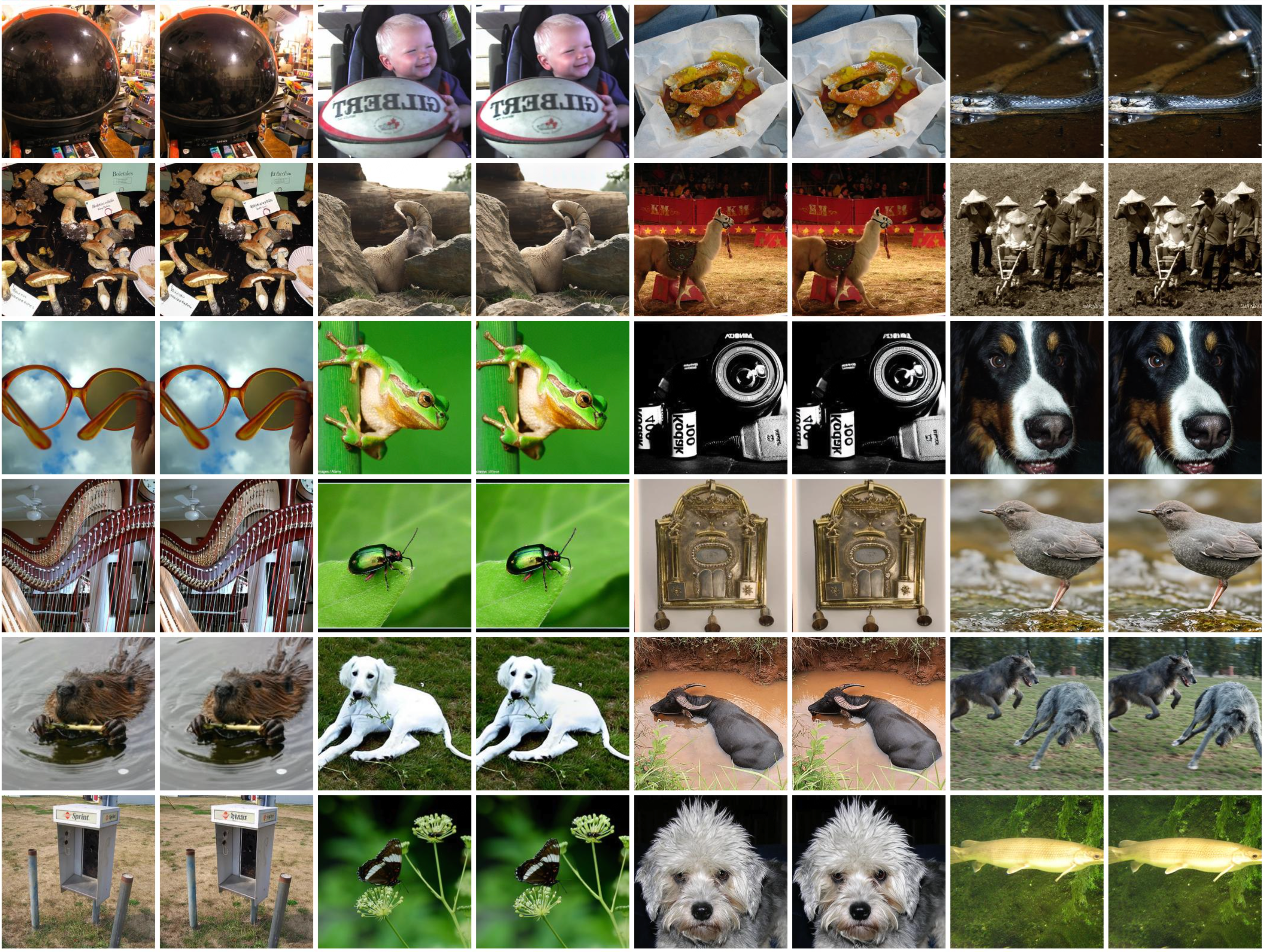} \\
\includegraphics[width=0.95\linewidth]{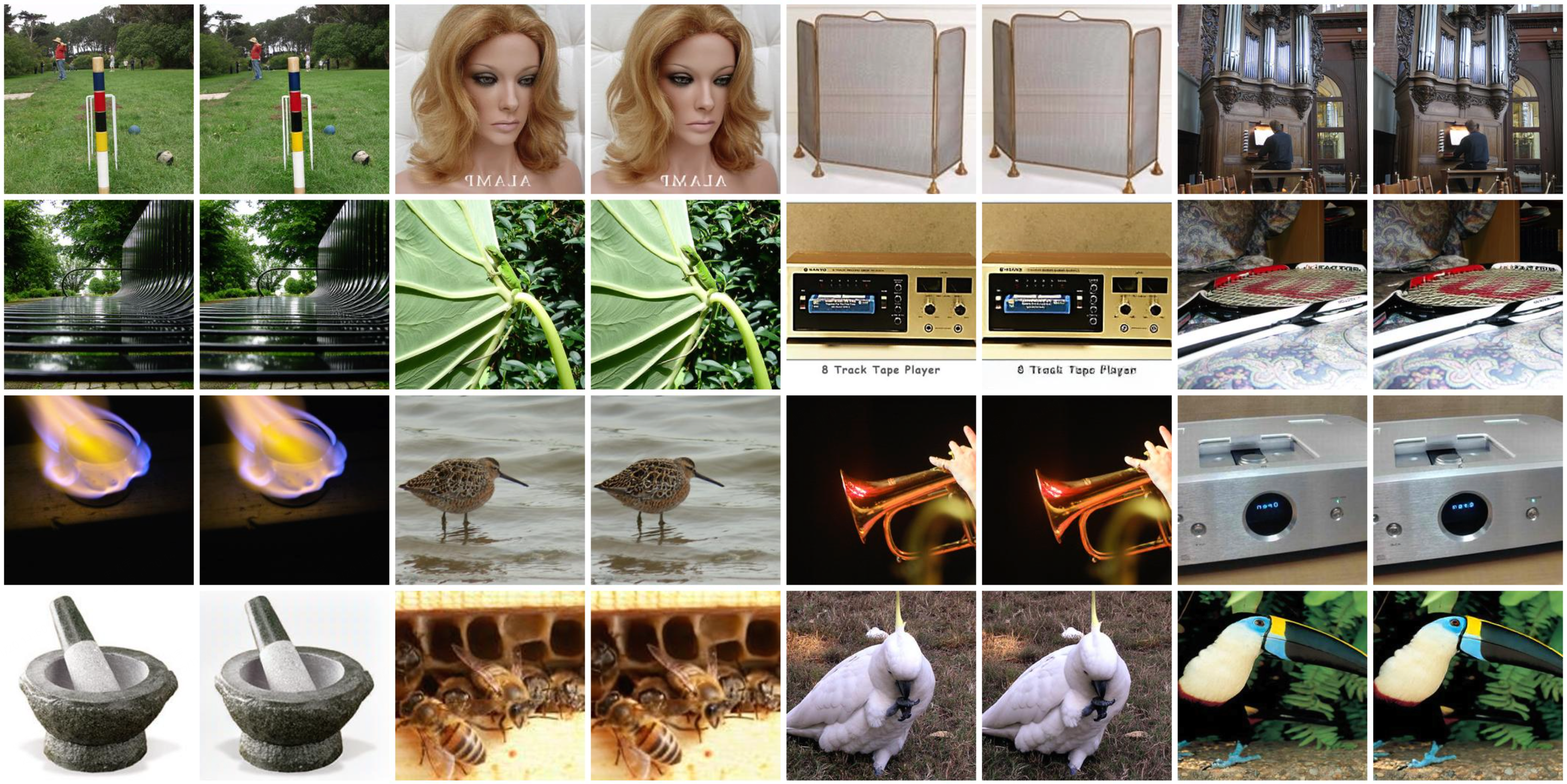}

\caption{\textbf{Supplementary reconstruction (1/1).} We put the original images on the left and the reconstructed images on the right.}
\label{fig:supp_rec_1}
\end{figure*}

\noindent
\textbf{Reconstruction Results.}
In Fig.~\ref{fig:supp_rec_1}, we present additional reconstruction results. RecTok accurately preserves the structure, color, and fine details of the input images.

\begin{figure*}[p]
\centering

\includegraphics[width=0.24\linewidth]{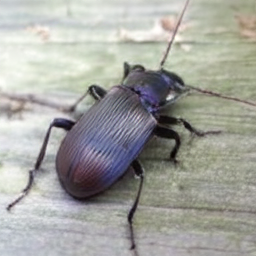}
\includegraphics[width=0.24\linewidth]{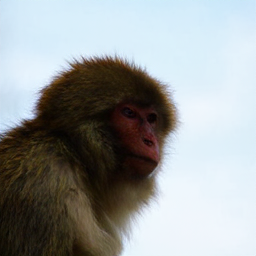}
\includegraphics[width=0.24\linewidth]{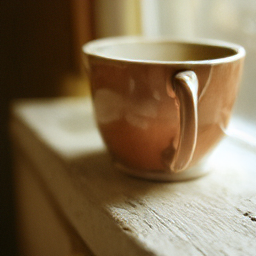}
\includegraphics[width=0.24\linewidth]{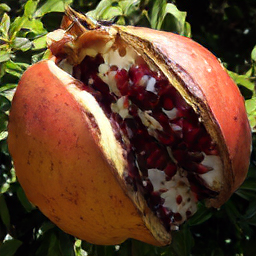}\\[2pt]

\includegraphics[width=0.24\linewidth]{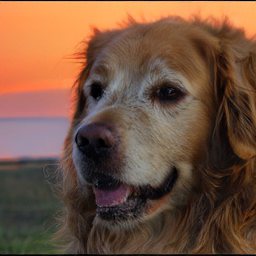}
\includegraphics[width=0.24\linewidth]{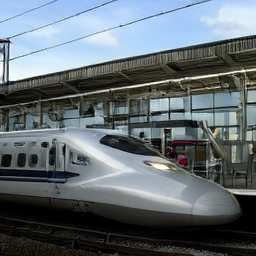}
\includegraphics[width=0.24\linewidth]{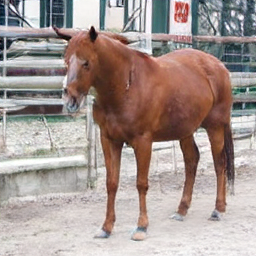}
\includegraphics[width=0.24\linewidth]{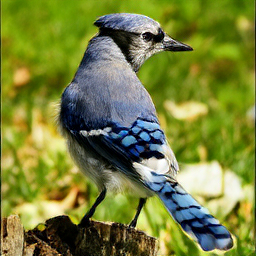}\\[2pt]

\includegraphics[width=0.24\linewidth]{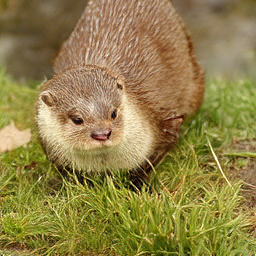}
\includegraphics[width=0.24\linewidth]{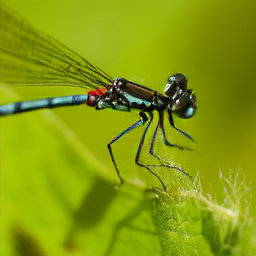}
\includegraphics[width=0.24\linewidth]{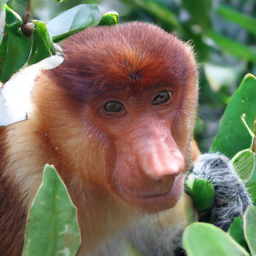}
\includegraphics[width=0.24\linewidth]{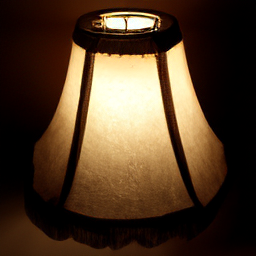}\\[2pt]

\includegraphics[width=0.24\linewidth]{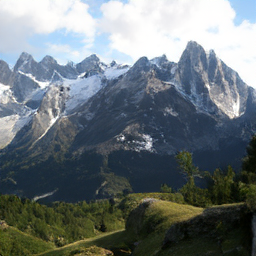}
\includegraphics[width=0.24\linewidth]{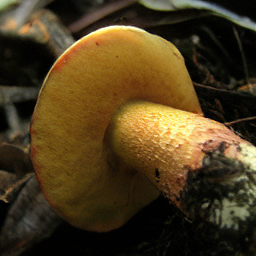}
\includegraphics[width=0.24\linewidth]{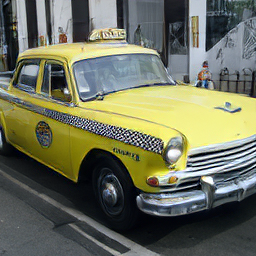}
\includegraphics[width=0.24\linewidth]{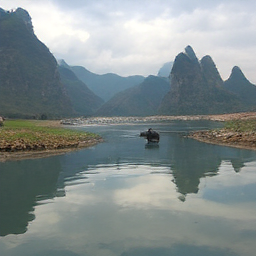}\\[2pt]

\includegraphics[width=0.24\linewidth]{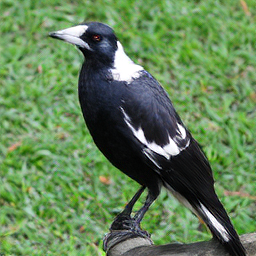}
\includegraphics[width=0.24\linewidth]{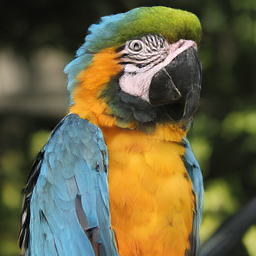}
\includegraphics[width=0.24\linewidth]{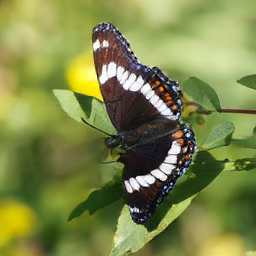}
\includegraphics[width=0.24\linewidth]{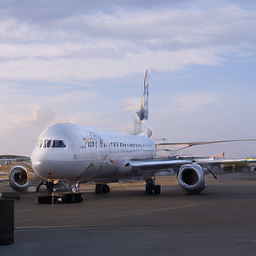}\\[2pt]

\caption{\textbf{Supplementary generations (1/3).}}
\label{fig:supp_gen_1}
\end{figure*}

\begin{figure*}[p]
\centering

\includegraphics[width=0.24\linewidth]{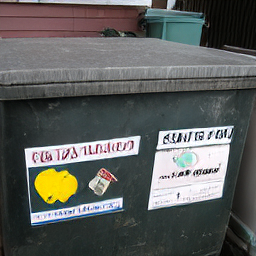}
\includegraphics[width=0.24\linewidth]{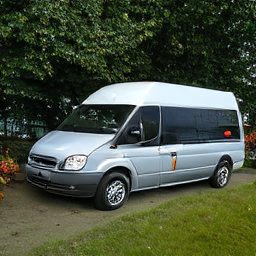}
\includegraphics[width=0.24\linewidth]{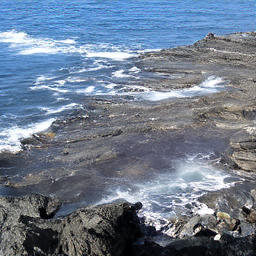}
\includegraphics[width=0.24\linewidth]{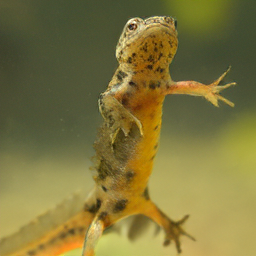}\\[2pt]

\includegraphics[width=0.24\linewidth]{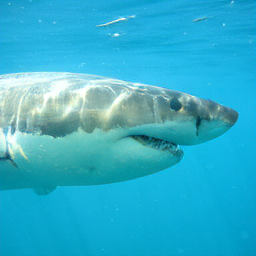}
\includegraphics[width=0.24\linewidth]{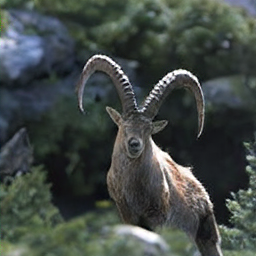}
\includegraphics[width=0.24\linewidth]{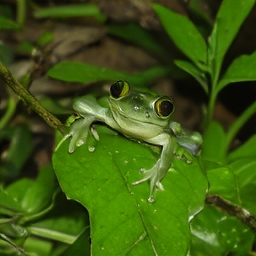}
\includegraphics[width=0.24\linewidth]{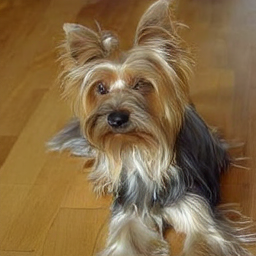}\\[2pt]

\includegraphics[width=0.24\linewidth]{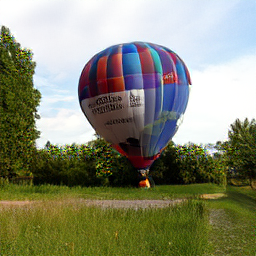}
\includegraphics[width=0.24\linewidth]{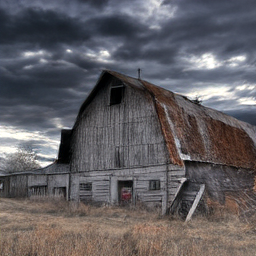}
\includegraphics[width=0.24\linewidth]{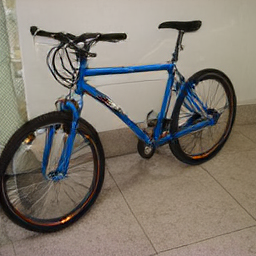}
\includegraphics[width=0.24\linewidth]{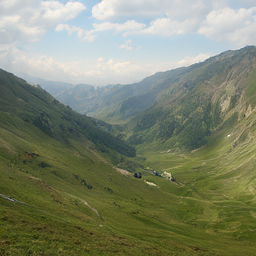}\\[2pt]

\includegraphics[width=0.24\linewidth]{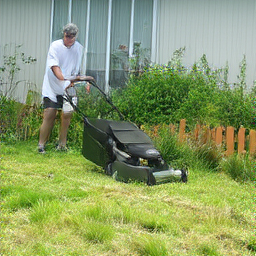}
\includegraphics[width=0.24\linewidth]{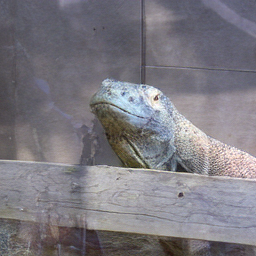}
\includegraphics[width=0.24\linewidth]{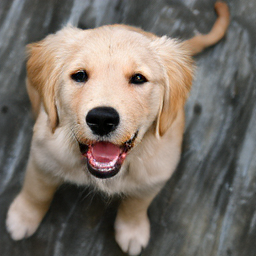}
\includegraphics[width=0.24\linewidth]{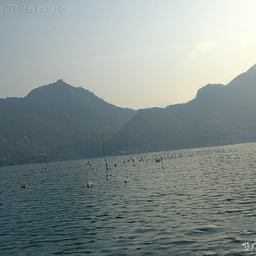}\\[2pt]

\includegraphics[width=0.24\linewidth]{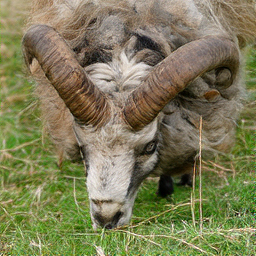}
\includegraphics[width=0.24\linewidth]{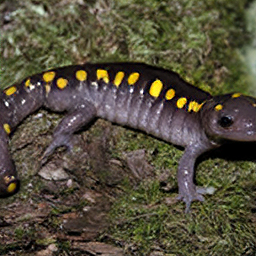}
\includegraphics[width=0.24\linewidth]{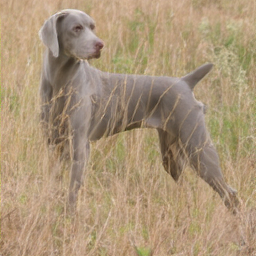}
\includegraphics[width=0.24\linewidth]{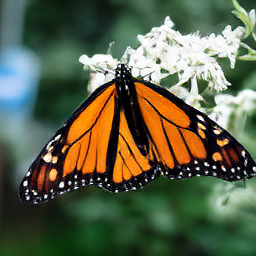}\\[2pt]

\caption{\textbf{Supplementary generations (2/3).}}
\label{fig:supp_gen_2}
\end{figure*}

\begin{figure*}[p]
\centering

\includegraphics[width=0.24\linewidth]{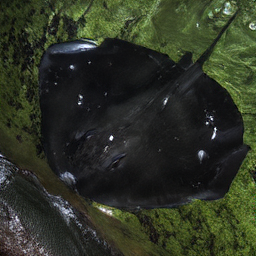}
\includegraphics[width=0.24\linewidth]{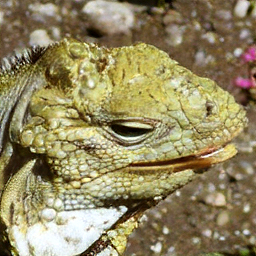}
\includegraphics[width=0.24\linewidth]{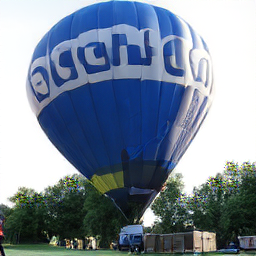}
\includegraphics[width=0.24\linewidth]{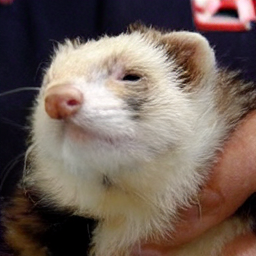}\\[2pt]

\includegraphics[width=0.24\linewidth]{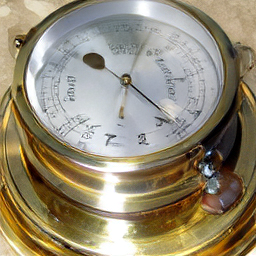}
\includegraphics[width=0.24\linewidth]{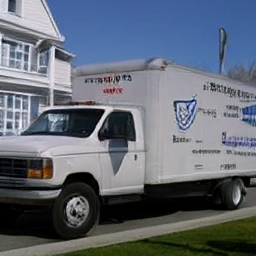}
\includegraphics[width=0.24\linewidth]{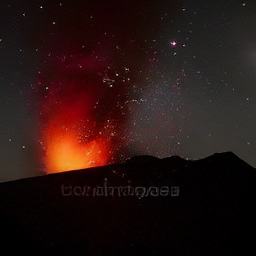}
\includegraphics[width=0.24\linewidth]{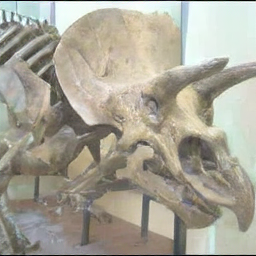}\\[2pt]

\includegraphics[width=0.24\linewidth]{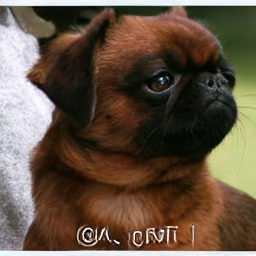}
\includegraphics[width=0.24\linewidth]{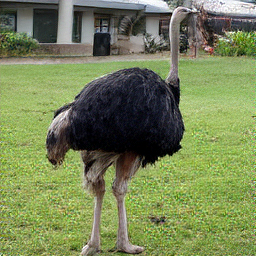}
\includegraphics[width=0.24\linewidth]{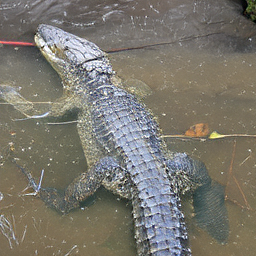}
\includegraphics[width=0.24\linewidth]{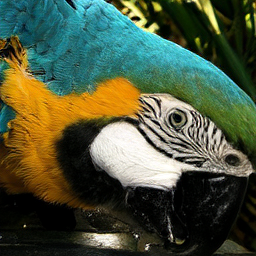}\\[2pt]

\includegraphics[width=0.24\linewidth]{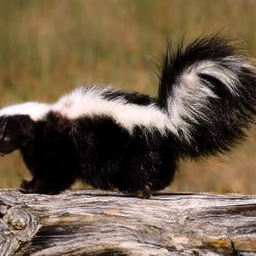}
\includegraphics[width=0.24\linewidth]{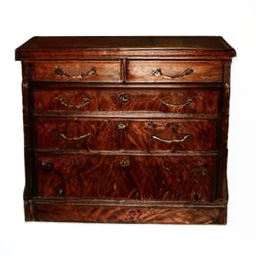}
\includegraphics[width=0.24\linewidth]{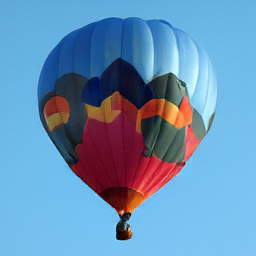}
\includegraphics[width=0.24\linewidth]{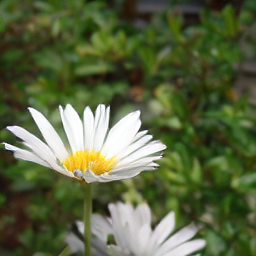}\\[2pt]

\includegraphics[width=0.24\linewidth]{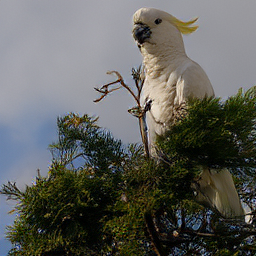}
\includegraphics[width=0.24\linewidth]{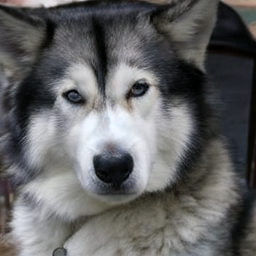}
\includegraphics[width=0.24\linewidth]{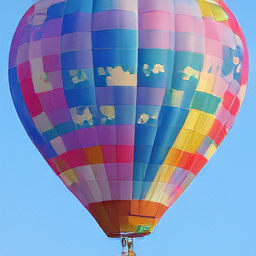}
\includegraphics[width=0.24\linewidth]{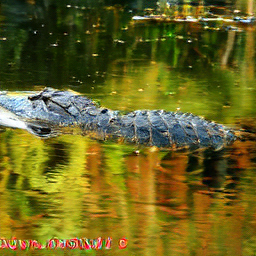}\\[2pt]

\caption{\textbf{Supplementary generations (3/3).}}
\label{fig:supp_gen_3}
\end{figure*}

\noindent
\textbf{Generation Results.}
In Fig.~\ref{fig:supp_gen_1}, Fig.~\ref{fig:supp_gen_2}, and Fig.~\ref{fig:supp_gen_3}. We provide additional generation results produced by a $\text{DiT}^{\text{DH}}-\text{XL}$ model trained for 600 epochs. We show outputs both with and without classifier-free guidance.

\section{Additional Analysis on RecTok}
\label{sec:additional_analysis_rectok}

\noindent
\textbf{The Discriminative Ability along the Flow.}
We visualize the features along the forward flow $x_t$ using t-SNE, uniformly sampling timesteps $t \in [0, 1]$. 
As shown in Fig.~\ref{fig:flow}, our RecTok exhibits strong semantic consistency throughout the forward flow. 
The t-SNE visualization and linear probing accuracy demonstrate the stable discriminative ability of $x_t$. 
A more discriminative $x_t$ also encourages forward trajectories to avoid intersections, which aligns with the objective of the original rectified flow~\cite{rf}.

\noindent
\textbf{Latent Feature Visualization.}
We visualize the latent features of RecTok using cosine similarity. 
Specifically, we extract the latent features using RecTok and obtain a global feature via spatial pooling. 
We then compute the cosine similarity between the global and latent features. 
%
%
We also show the PCA results, as shown in Fig.~\ref{fig:feat_cos} and Fig.~\ref{fig:feat_pca}, the heatmaps reveal that the learned latent features possess distinct semantic localization capabilities. Even without explicit segmentation supervision, the high similarity regions (indicated in warm colors) consistently align with the parts of the foreground objects, such as the head or the body structure. Conversely, background clutter and irrelevant textures are effectively suppressed. This suggests that RecTok’s tokenization process preserves spatial semantic integrity, ensuring that the aggregated global feature is highly representative of the core visual content and discriminative for downstream tasks.

\section{Limitations}
\label{sec:limitations}
In terms of semantics, although RecTok enhances semantic structure by increasing the latent dimensionality, its discriminative capability still lags behind that of VFMs. For example, DINOv3~\cite{dinov3}, SigLIP 2~\cite{siglip2}, and SAM~\cite{sam}.
Regarding reconstruction, while the KL loss smooths the latent space and improves generation quality, it inevitably weakens reconstruction ability, resulting in RecTok performing worse than an AE model with the same architecture. 
We leave these challenges as open questions for future work. We believe they can be addressed by further increasing the latent dimensionality and refining the KL regularization.

\section{Broader Impacts}
\label{sec:broader_impacts}
RecTok further expands the dimensionality of visual tokenizers while delivering consistent improvements in reconstruction, generation, and semantic representation. 
Its effectiveness suggests that the community may benefit from exploring higher-dimensional latent spaces that excel at both generative and understanding tasks~\cite{bai2024survey, mathscape, lovr, mmverify, synthvlm}. 
Such a latent space serves as a \textit{real unified representation}, removing the need for two separate image tokenizers as in prior unified models~\cite{wu2024janus}.
A shared feature space for both generation and understanding can promote mutual benefits across tasks, making unified models more coherent and meaningful.

\clearpage

\clearpage

\clearpage

\clearpage


\end{document}